\documentclass[acmtog, authorversion]{acmart}

\usepackage{booktabs} 
\usepackage[abs]{overpic}
\usepackage{subfiles}
\setcitestyle{square}

\usepackage[ruled]{algorithm2e} 

\graphicspath{
	{Figs/}
}

\SetAlFnt{\small}
\SetAlCapFnt{\small}
\SetAlCapNameFnt{\small}
\SetAlCapHSkip{0pt}
\IncMargin{-\parindent}

\newcommand{\R}{\mathbb{R}}
\let \bs=\mathbf
\let \set=\mathcal

\def \diag {\mathrm{diag}}

\def \saliency {\textup{\saliency}}

\def \init {\mathit{in}}
\def \path {\mathit{path}}

\acmJournal{TOG}
\acmVolume{37}
\acmNumber{6}
\acmArticle{1}
\acmYear{2018}
\acmMonth{7}

\setcopyright{usgovmixed}

\acmDOI{0000001.0000001_2}

\received{February 2007}
\received{March 2009}
\received[final version]{June 2009}
\received[accepted]{July 2009}

\begin{document}
\title{Deep Generative Modeling for Scene Synthesis via Hybrid Representations} 
\author{Zaiwei Zhang, Zhenpei Yang}
\orcid{1234-5678-9012-3456}
\affiliation{%
\institution{The University of Texas at Austin}
\streetaddress{2317 Speedway}
\city{Austin}
\state{TX}
\postcode{78712}
\country{USA}}
	
\renewcommand\shortauthors{Zhou, G. et al}
	
\begin{abstract}
We present a deep generative scene modeling technique for indoor environments. Our goal is to train a generative model using a feed-forward neural network that maps a prior distribution (e.g., a normal distribution) to the distribution of primary objects in indoor scenes. We introduce a 3D object arrangement representation that models the locations and orientations of objects, based on their size and shape attributes. Moreover, our scene representation is applicable for 3D objects with different multiplicities (repetition counts), selected from a database. We show a principled way to train this model by combining discriminator losses for both a 3D object arrangement representation and a 2D image-based representation. We demonstrate the effectiveness of our scene representation and the deep learning method on benchmark datasets. We also show the applications of this generative model in scene interpolation and scene completion. 
\end{abstract}

%
%

	
%
%

	
\author{Chongyang Ma, Linjie Luo}
\affiliation{%
\institution{Snapchat Inc.}}
\author{Alexander Huth, Etienne Vouga, Qixing Huang}
\affiliation{%
\institution{The University of Texas at Austin}
\streetaddress{2317 Speedway}
\city{Austin}
\state{TX}
\postcode{78712}
\country{USA}}
	
	
\begin{teaserfigure}
\centering 
\begin{tabular}{c|c|c|c|c|c}
\includegraphics[width=0.147\textwidth, trim=15px 30px 15px 0px, clip]{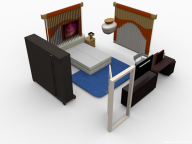}
&
\includegraphics[width=0.147\textwidth, trim=15px 30px 15px 0px, clip]{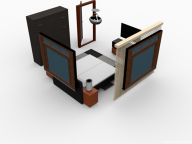}
&
\includegraphics[width=0.147\textwidth, trim=15px 30px 15px 0px, clip]{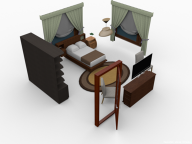}
&
\includegraphics[width=0.147\textwidth, trim=15px 30px 15px 0px, clip]{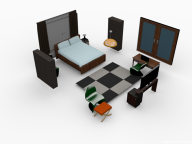}
&
\includegraphics[width=0.147\textwidth, trim=15px 30px 15px 0px, clip]{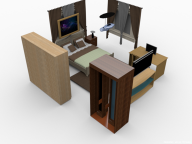}
&
\includegraphics[width=0.147\textwidth, trim=15px 30px 15px 0px, clip]{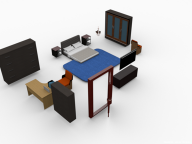}\\
\includegraphics[width=0.147\textwidth, trim=15px 30px 15px 0px, clip]{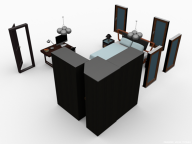}
&
\includegraphics[width=0.147\textwidth, trim=15px 30px 15px 0px, clip]{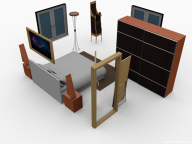}
&
\includegraphics[width=0.147\textwidth, trim=15px 30px 15px 0px, clip]{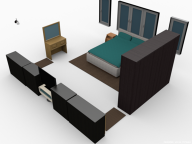}
&
\includegraphics[width=0.147\textwidth, trim=15px 30px 15px 0px, clip]{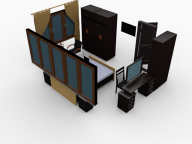}
&
\includegraphics[width=0.147\textwidth, trim=15px 30px 15px 0px, clip]{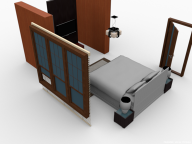}
&
\includegraphics[width=0.147\textwidth, trim=15px 30px 15px 0px, clip]{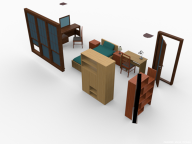}
\\ \hline
\includegraphics[width=0.147\textwidth, trim=15px 30px 15px 0px, clip]{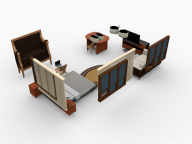}
&
\includegraphics[width=0.147\textwidth, trim=15px 30px 15px 0px, clip]{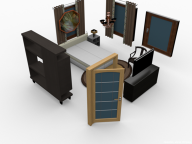}
&
\includegraphics[width=0.147\textwidth, trim=15px 30px 15px 0px, clip]{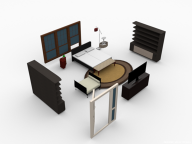}
&
\includegraphics[width=0.147\textwidth, trim=15px 30px 15px 0px, clip]{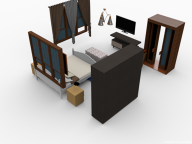}
&
\includegraphics[width=0.147\textwidth, trim=15px 30px 15px 0px, clip]{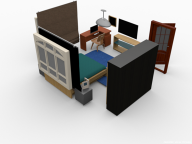}
&
\includegraphics[width=0.147\textwidth, trim=15px 30px 15px 0px, clip]{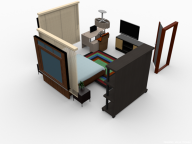}
\\
\includegraphics[width=0.147\textwidth, trim=15px 30px 15px 0px, clip]{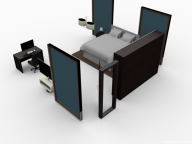}
&
\includegraphics[width=0.147\textwidth, trim=15px 30px 15px 0px, clip]{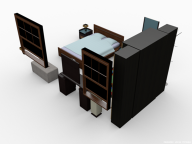}
&
\includegraphics[width=0.147\textwidth, trim=15px 30px 15px 0px, clip]{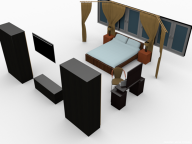}
&
\includegraphics[width=0.147\textwidth, trim=15px 30px 15px 0px, clip]{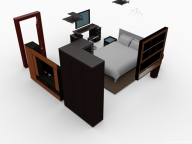}
&
\includegraphics[width=0.147\textwidth, trim=15px 30px 15px 0px, clip]{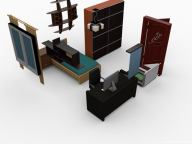}
&
\includegraphics[width=0.147\textwidth, trim=15px 30px 15px 0px, clip]{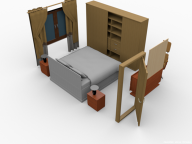}
\end{tabular}
\caption{Comparisons between our randomly generated scenes (row 1 and row 3) and their nearest neighbors in the training data (row 2 and row 4). Our synthesized scenes present significant topological and geometrical variations from the training data.}
\label{Figure:Best:Figure}
\end{teaserfigure}
	
\maketitle
	
\section{Introduction}
\label{Sec:Introduction}

Constructing 3D digital environments using low-dimensional parametric models is one the main tasks for scene generation in computer graphics. Such models enable a wide range of applications, including synthesizing new 3D models~\cite{3dgan}, constrained shape modeling and design~\cite{Yang:2011:SSE}, shape exploration and retrieval~\cite{akzm_shapeSynth_eg14,Schulz:2017:RPS}, and data-driven geometry reconstruction~\cite{Allen:2003:SHB,Anguelov:2005:SSC}. To date, however, most parametric models are either hand-crafted or learned from homogeneous data (e.g., faces~\cite{Blanz:1999:MMS} and human body models~\cite{Allen:2003:SHB,Anguelov:2005:SSC}). The remaining challenge still lies in learning effective parametric models from heterogeneous data that exhibits significant geometrical and structural variability. A typical and prominent example of this challenge is learning the space of 3D scenes, which consist of objects with large variations in category label, shape, pose, and repetition counts.

In this paper, we present an approach that uses a feed-forward neural network to parameterize the space of 3D indoor scenes from  a low-dimensional parametric vector. Our feed-forward architecture differs from  scene synthesis/modeling approaches\cite{Chaudhuri:2011:PRA,Fisher:2012:ESO,Chaudhuri:2013:ACC,Liu:2014:CCS,DBLP:journals/corr/IzadiniaSS16,Kermani:2016:LSS,Li:2017:GGR,Ritchie:2016:NPM,DBLP:journals/corr/HaE17,DBLP:journals/corr/abs-1712-08290,DBLP:conf/iccv/ZouYYCH17,Wang:2018:PRM}, which construct a 3D scene by iteratively adding new objects. Moreover, we are also unlike  prior neural network based 3D synthesis techniques that focus on volumetric grid representations~\cite{conf/cvpr/WuSKYZTX15,NIPS2016_6096,Wang:2017:OOC,DBLP:journals/corr/HaneTM17,DBLP:journals/corr/RieglerUBG17,DBLP:journals/corr/KlokovL17} or point cloud surface representations~\cite{DBLP:journals/corr/QiYSG17}. In our approach we represent a 3D scene as a configurational arrangement of objects, a representation that is favored by numerous 3D graphics representations such as scene editing to 3D augmented/virtual reality. 

In developing a feed-forward generative model using  this configurational representation, we overcome a number of technical challenges. Our approach combines three key ideas, ranging from methods to encode object arrangements in vectorized forms, determining a suitable generative model, and in designing effective procedures to learn the generative model from limited training data. To parameterize the 3D scene, we maintain a superset of abstract objects. Each scene is represented by selecting a subset of objects, and then determining the geometric shape, location, size and orientation of each object. We show that this encoding admits a natural matrix parameterization. Moreover, our encoding allows multiple objects in the same category (e.g., multiple chairs in the same scene). Second, we introduce a sparsely connected feed-forward neural network for generating this scene representation from a low-dimensional latent parameter. This network design adequately addresses the overfitting issue of using fully connected layers. Moreover, we train this generator using a combination of an arrangement autoencoder loss, an arrangement discriminator loss, and an image-based discriminator loss. In particular, the image-based discriminator loss is defined on the top-view rendered image of indoor scenes. This also effectively addresses local compatibility issues that arise when training under the arrangement representation alone. Note that although we use a hybrid representation for training, during testing time we only use the feed-forward generator for scene synthesis. 

We have applied our approach on synthesizing living rooms and bedrooms using the SUNCG dataset~\cite{song2016ssc}. The living and bedroom categories consist of 5688 and 5922  scenes respectively. For each category, we use 5000 scenes for training and leave the remaining scenes as testing. Our approach trains 3D scene generators with 1,198 minutes and 1,001 minutes, respectively, using a desktop with 3.2GHZ CPU, 64G main memory and a Titan X GPU. The trained generators can synthesize diverse and novel 3D scenes that are not present in the training sets (See Figure~\ref{Figure:Best:Figure}). Synthesizing one scene takes 30 ms. We present quantitative evaluations to justify our design choices. We also show the usefulness of the approach on applications of scene interpolation/extrapolation and scene completion. 

In summary, we present the following main contributions:
\begin{itemize}
\item We show that it is possible to train a feed-forward parametric generative model that maps a latent parameter to a 3D indoor scene. The 3D scene is represented as an arrangement of 3D objects, and each category of objects may repeat multiple times.
\item We introduce a methodology for 3D scene synthesis using hybrid representations, which combine a 3D object arrangement representation for capturing coarse object interactions and an image-based representation for capturing local object interactions. 
\end{itemize}

\section{Related Works}

\paragraph{Hand-crafted generative models.}
Early work in parametric shape modeling consists of shapes designed by domain experts. Examples include work for trees~\cite{Weber:1995:CRR} and Greek doric temples~\cite{teboul:tel-00628906}. 
It can be prohibitively difficult, however, for humans to model complicated object classes that exhibit significant geometric and/or topological variability. For this reason, parametric models (or procedural models) for many model classes (e.g., furniture shapes and scenes) do not exist. 

\paragraph{Learning generative models.}
To faithfully capture shape variability in geometric data, a recent focus in visual computing is to learn parametric models from data. This trend is aligned with the significant growth of visual data available from online resources such as ImageNet~\cite{imagenet_cvpr09} and ShapeNet~\cite{DBLP:journals/corr/ChangFGHHLSSSSX15}. Methods for learning parametric models differ from each other in terms of the representation of the visual data as well as the mapping function. Early works on learning parametric models focus on Faces and Human bodies~\cite{Blanz:1999:MMS,Allen:2003:SHB,Anguelov:2005:SSC}, which can be parametrized by deforming a template model. The parametric models are given by linearly blending exemplar models in a model collection. Such a method is only applicable to object classes with small geometric variability and no topological variability. They are not suitable for indoor scenes that can exhibit significant topological and geometrical variability. 

Motivated from the tremendous success of deep neural networks, a recent focus has been on encoding the mapping function using neural networks. In the 2D image domain, people have developed successful methods for deep generative models such as generative adversarial networks (GANs)~\cite{NIPS2014_5423,DBLP:journals/corr/SalimansGZCRC16,DBLP:journals/corr/ZhaoML16,DBLP:journals/corr/ArjovskyCB17}, variational autoencoders~\cite{DBLP:journals/corr/KingmaW13,DBLP:journals/corr/KingmaSW16}, and autoregression~\cite{VanDenOord:2016:PRN}. Although these approaches work well on 2D images, extending them to 3D data is highly non-trivial. A particular challenge is to develop a suitable representation for 3D data. 

\paragraph{3D representations.}
Unlike other modalities that naturally admit vectorized representations (e.g., images and videos), there exists great flexibility when encoding 3D geometry in their vectorized forms. In the literature, people have developed neural networks for multi-view representations~\cite{Su:2015:MCN,DBLP:conf/cvpr/QiSNDYG16,DBLP:conf/eccv/TatarchenkoDB16}, volumetric representations~\cite{conf/cvpr/WuSKYZTX15,NIPS2016_6096,Wang:2017:OOC,DBLP:journals/corr/HaneTM17,DBLP:journals/corr/RieglerUBG17,DBLP:journals/corr/KlokovL17,DBLP:journals/corr/abs-1712-01812}, point-based representations~\cite{DBLP:conf/cvpr/QiSMG17}, part-based representations~\cite{DBLP:conf/cvpr/TulsianiSGEM17,Li:2017:GGR}, graph/mesh representations~\cite{Masci:2015:GCN,DBLP:journals/corr/HenaffBL15,DBLP:journals/corr/YiSGG16,DBLP:conf/cvpr/MontiBMRSB17} and spherical representations~\cite{DBLP:journals/corr/abs-1801-10130,DBLP:journals/corr/abs-1711-06721,DBLP:journals/corr/abs-1712-04426}. 

Building parametric 3D models have mostly focused on 3D shapes. Wu et al.~\cite{NIPS2016_6096} describe a 3D generative network under the volumetric representation. Extending this approach to 3D scenes faces the fundamental challenge of limited resolution. In addition, its output is not an arrangement of objects. \cite{DBLP:conf/cvpr/TulsianiSGEM17} proposed a part-based model for synthesizing 3D shapes that are described as an arrangement of parts. \cite{Nash:2017:SVA} proposed ShapeVAE for synthesizing 3D shapes that are described as a semantically labeled point cloud. Both approaches are specifically tailored for 3D shapes, and it is challenging to extend them to 3D scenes. For example, both approaches require that shapes are consistently oriented, and such orientations are not available for 3D scenes. In our approach, we jointly optimize both the generators and the orientations of the input scenes. In addition, we found that variations in 3D scenes are more significant than 3D shapes, and approaches which work well on shapes generally lead to sub-optimal results on 3D scenes, e.g., spatial relations between correlated objects are not captured well. This motivates us to develop new representations and training methods for 3D scenes. 

The difference between our approach and existing 3D synthesis approaches is that we combine training losses under two representations, i.e., an object arrangement representation and an image-based representation. This innovative design allows us to obtain globally meaningful and locally compatible synthesis results. 

\paragraph{Assembly-based geometric synthesis.} Currently the dominant 3D scene synthesis method is assembly-based. Funkhouser et al. \cite{Funkhouser:2004:ME} introduced the first system that generates new 3D models by assembling parts from existing models. People have also applied this concept for various applications such as interactive modeling~\cite{conf/pg/KraevoyJS07}, design
\cite{Chaudhuri:2010:DSC}, reconstruction
~\cite{Shen:2012:SRP,Huang:2015:SRV}, and synthesis~\cite{Xu:2012:FDS}. The advantage of these methods is that they can handle datasets with significant structural variability. The downside is that these methods require complicated systems and careful parameter tuning. To improve system performance, a recent line of works utilize probabilistic graphical models (e.g., Bayesian networks) for assembly-based modeling and synthesis ~\cite{Merrell:2010:CRB,Chaudhuri:2011:PRA,Kalogerakis:2012:PMC,Fisher:2012:ESO,Chaudhuri:2013:ACC,Xu13sig,Chen:2014:ASM,Liu:2014:CCS,DBLP:journals/corr/IzadiniaSS16,Kermani:2016:LSS,Sung:2017:CWC}. Along this line, several works~\cite{DBLP:conf/icml/JiangLS12,Fisher:2015:ASS,Savva:2016:PLI,Ma:2016:AIS,Qi:2018:HIS} focus on using human interactions with objects and/or human actions to guide the synthesis process. These methods significantly stabilize the modeling and synthesis process. The nodes and edges in the graphical models, however, are usually pre-defined, which necessitates significant domain knowledge. 

Another recent line of works~\cite{Li:2017:GGR,Ritchie:2016:NPM,DBLP:journals/corr/HaE17,DBLP:journals/corr/abs-1712-08290,DBLP:conf/iccv/ZouYYCH17,Wang:2018:PRM} reformulate assembly-based synthesis as recursive procedures. Starting from a root part, these methods recursively insert new parts conditioned on existing parts. This conditional probability is described as a neural network. In contrast, our approach proposes to learn 3D synthesis using a feed-forward network. In particular, our approach does not require hierarchical labels (either provided by users or generated computationally) that are required for training such recursive procedures. 

Priors learned from training data can be used for rectifying 3D scenes as well. \cite{Yu:2011:MHA} present an optimization framework for turning a coarse object arrangement into significantly improved object arrangements. Our image-based discriminator loss is conceptually similar to this approach, yet we automatically learn this loss term from data.

\paragraph{Image-based representation for 3D synthesis}

Several recent works leverage image-based representations for 3D synthesis. In ~\cite{DBLP:journals/corr/abs-1803-08999}, the authors leverage the image-based representation to predict the locations of key objects in a scene. In ~\cite{Wang:2018:PRM}, the authors use an image-based representation to predict locations and other attributes of the object to be inserted. In contrast, our approach learns a parametric 3D generator for synthesis. The image-based representation, which serves as a regularizer for the 3D generator, is only used in the training process. 

\section{Problem Statement and Approach Overview}
\label{Section:Overview}
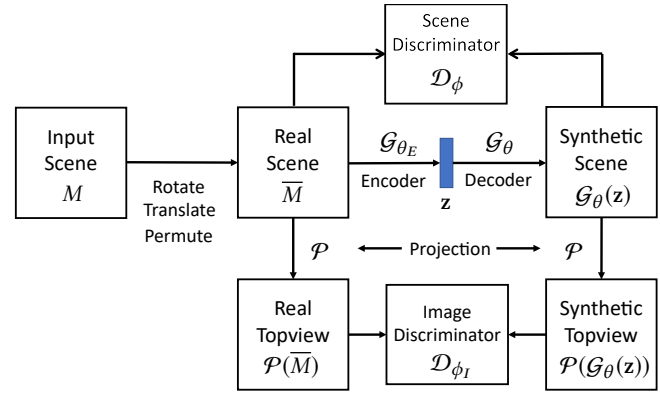
\begin{figure}
\centering
\begin{overpic}[width=1.0\columnwidth]{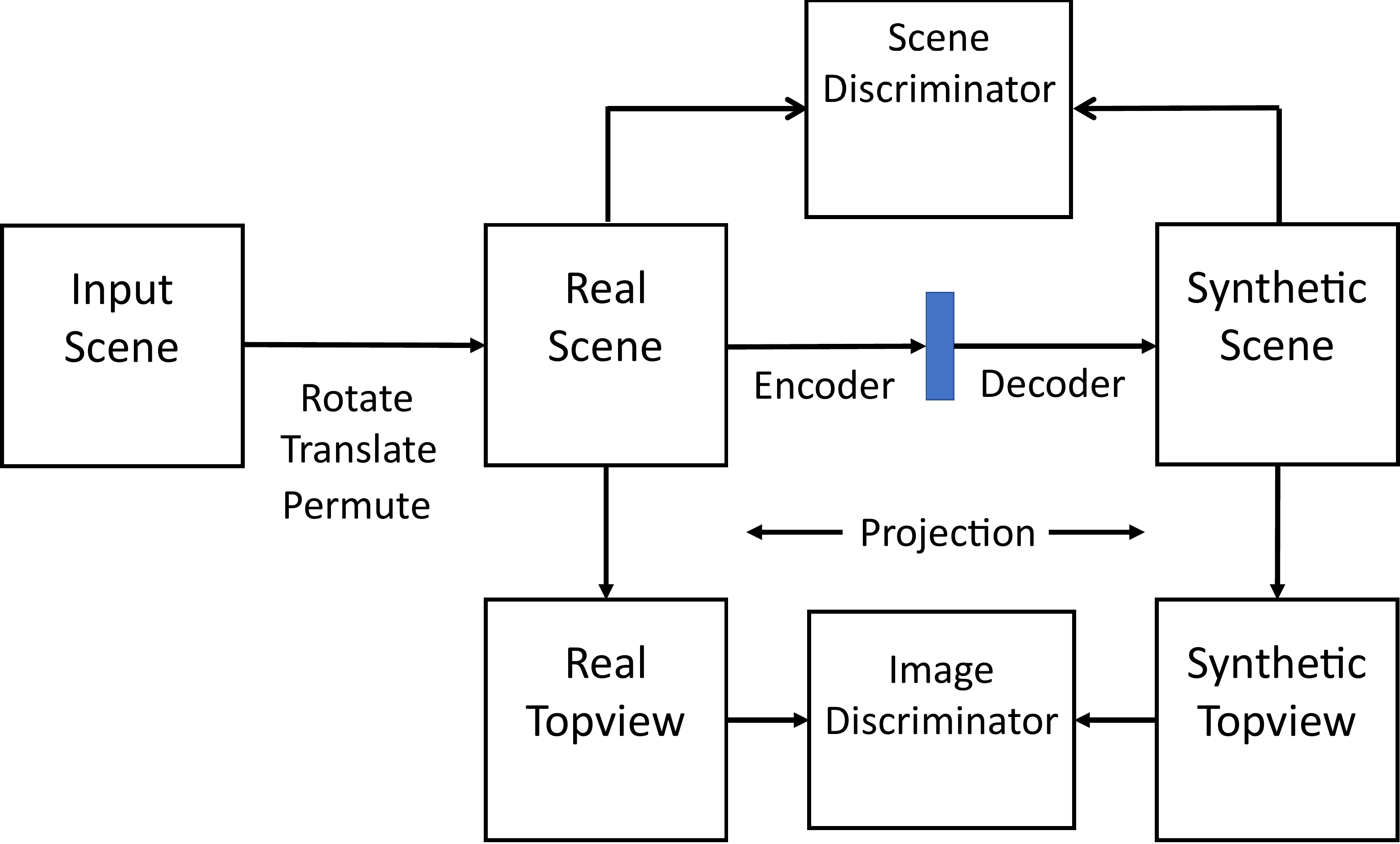}
\put(212,71.5){$\set{G}_{\theta}(\bs{z})$}
\put(101,71.5){$\overline{M}$}
\put(18,72){$M$}
\put(206,07){$\set{P}(\set{G}_{\theta}(\bs{z}))$}
\put(157,08){$\set{D}_{\phi_I}$}
\put(157,116){$\set{D}_{\phi}$}
\put(94,07){$\set{P}(\overline{M})$}
\put(112,50){$\set{P}$}
\put(208,50){$\set{P}$}
\put(161,69){$\bs{z}$}
\put(138,92){$\set{G}_{\theta_E}$}
\put(178,92){$\set{G}_{\theta}$}
\end{overpic}
\vspace{0.03in} 
\caption{Illustration of different modules of our design.}
\label{Figure:Overview}
\end{figure}

In this section, we give an overview of the 3D scene synthesis problem  (Section~\ref{Subsection:Problem:Statement}) and of the proposed approach for solving it (Section~\ref{Subsection:Approach:Overview}).

\subsection{Problem Statement}
\label{Subsection:Problem:Statement}

Our goal is to train a neural network that takes a random variable $\bs{z}\in \R^d$ as input and generates a 3D scene. A scene consists of some number of rigid objects arranged in space in a semantically meaningful way, and free of interpenetrations. We assume each object belongs to one of a predefined set of object classes, and that objects within each class can be parameterized by a shape descriptor (this descriptor is then used to retrieve the object's 3D geometry from a shape database). We further assume that objects rest on the ground, and are correctly oriented with respect to the vertical direction, so that each object's placement in the scene can be specified by an orientation and position in the $xy$ plane (the \emph{top view}), as well as a nonuniform scaling. We formalize the scene representation in section~\ref{Subsection:Scene:Representation}.


To train our networks, we use $N$ different 3D scenes $S_1,\cdots, S_{N}$ gathered from 3D Warehouse\footnote{https://3dwarehouse.sketchup.com/}. We do not require that the scenes are globally oriented in a consistent way, that objects are specified in any particular order, etc; our training formulation is robust to such variations. In addition, our approach does not require additional local or global supervision such as a hierarchical grouping of objects in a scene. 

\subsection{Approach Overview}
\label{Subsection:Approach:Overview}
The central challenge of scene synthesis is that on the one hand, rigid arrangement of objects, and semantically-meaningful placement of objects relative to each other, is best captured by a vectorized representation as described in the previous section. On the other hand, such an abstract representation of scenes neglects geometric detail; for example, it is very difficult to compute, or learn, when a pair of transformations of two objects causes them to intersect.

To attack this problem, we make the following observation: in a typical indoor scene, objects have a well-defined ``up'' direction and are placed (either on the ground, or on other objects, or hanging from the ceiling) in a manner that preserves this direction. Moreover, a 2D image of the top view is usually sufficient for observing and evaluating the local geometric relationships between nearby objects.

We thus propose a design adapted to the dual global and local nature of the scene synthesis problem (see Figure~\ref{Figure:Overview}). We train a scene generator network by combining a variational autoencoder loss, which ensures coverage of the training data; a scene discriminator loss, which operates directly on a vectorized representation of the 3D scene and captures semantic patterns in the arrangement of scene objects; and an image discriminator loss operating on an image representing the 2D top view of the scene, which precisely captures the local geometric arrangement of nearby objects.


The reminder of this Section summarizes the design and motivation of each component of our design; Section~\ref{Sec:Approach} will spell out the technical details.

\paragraph{Object arrangement scene representation.}
We represent a 3D arrangement using a matrix whose columns correspond to the objects in the scene. In other words, each 3D scene is specified by selecting some number of objects of each object class and then arranging/fixing them in 3D space. Each column of the matrix representation describes the status of the corresponding object, namely, whether it appears in the scene or not, its location, size, orientation and shape. Notice that while each matrix completely specifies a 3D scene, this representation is redundant. To handle the technical challenge of non-uniqueness of this encoding (i.e., shuffling columns of the same category leads to the same scenes), we introduce latent permutation variables which effectively factor out such permutation variability. 

\paragraph{Scene generator.}
We design the scene generator as a feed-forward network with interleaved sparsely and fully connected layers operating on the matrix representation of scenes. The motivation for this architecture comes from the observations that (1) correlations among objects in an indoor scene are usually of low-order, and (2) sparsely connected layers have significantly reduced model size compared to fully connected networks, which improves generalization error.

\paragraph{Image-based module.}
We leverage a CNN-based discriminator loss, which captures the object correlations based on local object geometry that cannot be effectively captured by the matrix-encoding-based discriminator loss. Specifically, we encode each 3D scene as a 2D projection of the scenes objects onto the $xy$ plane. We impose a CNN-based image discriminator loss on this 2D image, which is back propagated to the scene generator, forcing it more accurately learn local correlation patterns.  

\paragraph{Joint scene alignment.}
Despite our fairly intuitive network design, training the network from unorganized scene collections is difficult. One challenge is that the training scenes are not necessarily globally oriented in a consistent way, nor do objects have consistent absolute locations. Moreover, although objects can be grouped by class, there is no canonical ordering of objects within the same class. To address these issues, we solve a global optimization problem to jointly align the input scenes in a training preprocessing step. We found that first aligning the input scenes significantly improves the resulting 3D scene generator. 

\paragraph{Training.}
Given roughly aligned input scenes, we learn the generator by optimizing an objective function that combines an autoencoder loss and the two discriminator losses described above. The variables include the generator network, two discriminators, the pose of each scene, and the orderings of the objects in each 3D scene. To facilitate the optimization, we introduce a latent variable for the scene that characterizes its underlying configuration (i.e., after transformation and object re-ordering). Both the autoencoder loss and discriminator losses are defined on this latent variable. In addition, we penalize the difference between this latent scene and its corresponding input scene (after transformation and object re-ordering). In doing so, the optimization variables are nicely decoupled, allowing efficient optimization via alternating minimization. 

\section{Approach}
\label{Sec:Approach}

In this section, we present technical details of our approach. In Section~\ref{Subsection:Scene:Representation}, we present a matrix representation of 3D object arrangement. In Section~\ref{Subsection:Scene:Generator} and Section~\ref{Subsection:Image:Based:Discriminator}, we describe the 3D object arrangement module and the image-based module, respectively. In Section~\ref{Subsection:Scene:Alignment}, we introduce how to jointly align the input scenes. Finally, we describe network training procedure in Section~\ref{Subsection:Network:Training}.

\subsection{Scene Representation}
\label{Subsection:Scene:Representation}

\begin{figure}
\centering
\begin{overpic}[width=1.0\columnwidth]{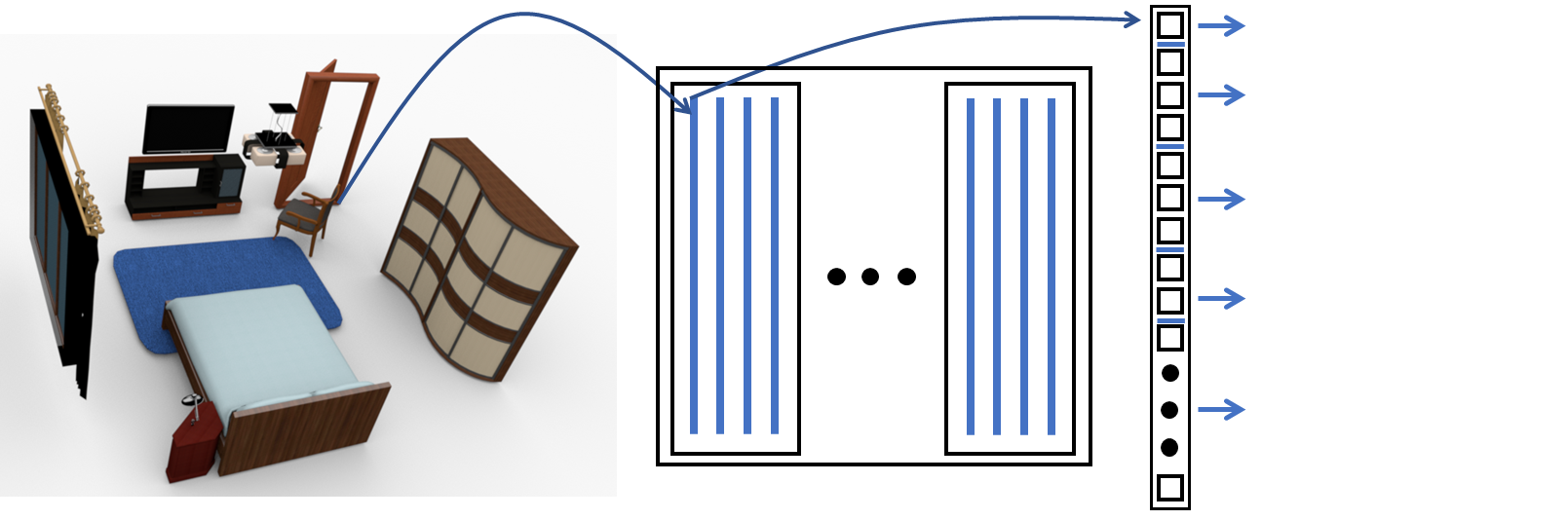}
\put(200,76){Existence}
\put(200,64){Location}
\put(200,48){Size}
\put(200,32){Orientation}
\put(200,15){Descriptor}
\put(45,-2){$S$}
\put(130,-2){$M$}
\put(178.5,-6){$\bs{v}$}
\end{overpic}
\vspace{0.03in}
\caption{Illustration of our scene encoding scheme.}
\label{Figure:Scene:Encoding}
\end{figure}
\begin{figure*}[t]
\centering
\begin{overpic}[width=1.0\textwidth]{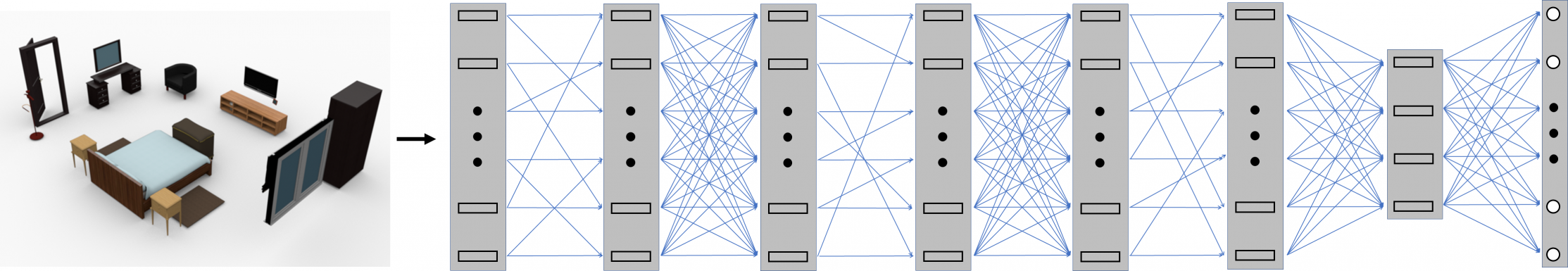}
\put(136,-14){\footnotesize{$120\times (d+9)$}}
\put(192,-14){\footnotesize{$200\times 10$}}
\put(246,-14){\footnotesize{$20\times 10$}}
\put(297,-14){\footnotesize{$80\times 20$}}
\put(348,-14){\footnotesize{$10\times 20$}}
\put(399,-14){\footnotesize{$20\times 20$}}
\put(451,-14){\footnotesize{$4\times 20$}}
\put(503,-14){\footnotesize{$\bs{z}$}}
\put(174,-4){SC}
\put(229,-4){FC}
\put(281,-4){SC}
\put(331,-4){FC}
\put(380,-4){SC}
\put(432,2){FC}
\put(480,2){FC}
\end{overpic}
\vspace{0.03in}
\caption{This figures illustrates the network module that is used for the encoder $\set{G}_{\theta_E}$. The decoder $\set{G}_{\theta}$ is reversed from the encoder $\set{G}_{\theta_E}$. The arrangement discriminator $\set{D}_{\phi}$ shares the same network architecture but replaces the latent vector by a value. This network module interweaves between sparsely connected (or SC) layers and fully connected (or FC) layers.}
\label{Figure:Network:Architecture}
\end{figure*}

To parameterize 3D scenes, we assume each object belongs to one of $n_c$ object categories, and that scenes can contain up to $m_k$ objects of each class $k$. Each scene therefore contains a maximum of $n_o = \sum_{k=1}^{n_c} m_k$ objects $\set{O}$. In our experiments, we use $n_c = 30$ and $m_k = 4, 1\leq k \leq n_c$ (see Section~\ref{Section:Experimental:Evaluation} for details). Note that another alternative encoding is to allow $n_o$ total number of arbitrary objects (c.f.~\cite{DBLP:journals/corr/TulsianiSGEM16,DBLP:journals/corr/FanSG16}). However, we found that explicitly encoding the class label of each object is superior to synthesizing the class label of each object, particularly when the number of distinctive classes is large.  


We assume that objects in each class can be uniquely identified with a $d$-dimensional shape descriptor, with $d$ constant across all classes. We can thus encode each object $o\in \set{O}$ using a status vector $\bs{v}^o \in \R^{d+9}$:
\begin{itemize}
\item $v_0^o$ is a tag that specifies whether $o$ appears in the scene ($v_0^o \geq 0.5$) or not ($v_0^o < 0.5$).
\item $(v_1^o, v_2^o, v_3^o)^{T}$ specifies the center of the bounding box of $o$ in a world coordinate system; we assume the up orientation of each object is always along the $z$-axis of this world coordinate system.
\item $(v_4^o, v_5^o)^{T}$ specifies the front-facing orientation of the bounding box of $o$ in the top view ($xy$ plane).
\item $v_6^o$, $v_7^o$, and $v_8^o$ specify the size of the bounding box of $o$ in the front, side, and up directions, respectively. 
\item $(v_9^o,\cdots, v_{d+8}^o)^{T}$ is the aforementioned descriptor that characterizes the geometric shape of $c$. In this paper, we use as our descriptor the second-to-last layer of the volumetric module of Qi et al.~ \shortcite{DBLP:conf/cvpr/QiSNDYG16} pre-trained on ShapeNetCore~\cite{DBLP:journals/corr/ChangFGHHLSSSSX15}. 
\end{itemize}
Given this object representation, a 3D scene can be parameterized by a matrix $M \in \R^{(d+9)\times n_o}$, with blocks of columns $M_k \in R^{(d+9)\times m_k}$ containing the status vectors of the objects of the $k$-th category.

One technical challenge of this intuitive encoding of 3D scenes is that it is invariant to global rigid motions, and to permutations of columns of the $M_k$. Both invariants need to be factored out to make the scene encoding unique. To that end, we introduce two operators on the matrix encoding $M$: the first operator 
\begin{align*}
    \set{S}(M; \sigma_1, \ldots, \sigma_{n_k}):\qquad&\quad\\ 
    \R^{(d+9)\times n_o} \times \prod_{k=1}^{n_c} S_{m_k} &\to \R^{(d+9)\times n_o}\\
\left[\begin{array}{ccc} M_1 & \cdots & M_{n_c}\end{array}\right] &\mapsto \left[\begin{array}{ccc} M_1\sigma_1 & \cdots & M_{n_c}\sigma_{n_c}\end{array}\right]
\end{align*}
applies column permutations $\sigma_k$ to objects of each class. To avoid clutter, in what follows we will elide the explicit dependence of $\set{S}$ on the $\sigma$. Note that $\set{S}$ applies permutations to objects of each class independently. Although introducing latent permutation variables is a common and effective technique~\cite{DBLP:journals/corr/TulsianiSGEM16,DBLP:journals/corr/FanSG16}, we could have employed permutation invariant functions instead~\cite{DBLP:conf/cvpr/QiSMG17,DBLP:journals/corr/QiYSG17}. However, we found that using permutation variables gives us more freedom to use powerful neural network layers. 

The second operator $\set{T}(M; R,\bs{t})$ applies a rotation $R\in SO(2)$ about the $z$ axis, and an arbitrary translation $\bs{t}\in \mathbb{R}^3$, to the bounding box position encoded within each column of $M$, and likewise applies $R$ to each object orientation. 
Again, we will elide $R$ and $\bs{t}$ when convenient. 

We associated a set of latent variables $\{\sigma_k, R, \bs{s}\}$ to each object $i$, and to each pair of objects $ij$; we will use the notation $\set{S}_i$, $\set{T}_{ij},$ etc, to interchangeably denote the operators, and the latent variables that determine them.

We factor out permutations of objects and the global pose of each input scene by introducing a latent matrix encoding $\overline{M}_i\in \R^{(d+9)\times n_o}$ for each object $i$. The antoencoder and discriminator losses described below will be imposed on $\overline{M}_i$, and we enforce consistency of $\overline{M}_i$ and $M_i$ by minimizing the loss
\begin{equation}
f_{d}\left(\overline{M}_i, M_i\right) = \min_{\set{T}_i, \set{S}_i}\left\|\overline{M}_i - \left(\set{T}_i \circ \set{S}_i\right)(M_i)\right\|_F^2,
\label{Eq:Scene:Distance}
\end{equation}
where $\|\cdot\|_F$ denotes the matrix Frobenius norm. Initialization and optimization of the latent permutation and rigid transformation variables are described below in section~\ref{Subsection:Scene:Alignment}.

\subsection{3D Object Arrangement Module}
\label{Subsection:Scene:Generator}
\begin{figure}
\includegraphics[width=0.15\textwidth, trim=15px 30px 15px 0px, clip]{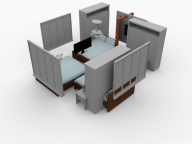}
\includegraphics[width=0.15\textwidth, trim=15px 30px 15px 0px, clip]{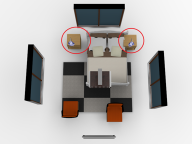}
\includegraphics[width=0.15\textwidth, trim=15px 30px 15px 0px, clip]{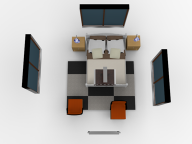}
\includegraphics[width=0.15\textwidth, trim=15px 30px 15px 0px, clip]{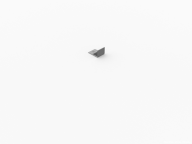}
\includegraphics[width=0.15\textwidth, trim=15px 30px 15px 0px, clip]{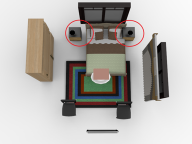}
\includegraphics[width=0.15\textwidth, trim=15px 30px 15px 0px, clip]{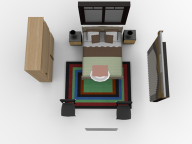}
\includegraphics[width=0.15\textwidth, trim=15px 30px 15px 0px, clip]{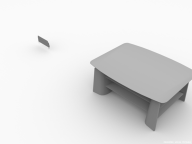}
\includegraphics[width=0.15\textwidth, trim=15px 30px 15px 0px, clip]{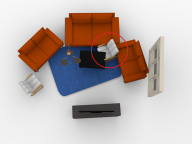}
\includegraphics[width=0.15\textwidth, trim=15px 30px 15px 0px, clip]{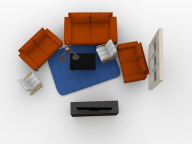}
\includegraphics[width=0.15\textwidth, trim=15px 30px 15px 0px, clip]{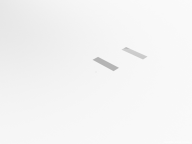}
\includegraphics[width=0.15\textwidth, trim=15px 30px 15px 0px, clip]{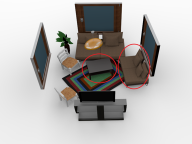}
\includegraphics[width=0.15\textwidth, trim=15px 30px 15px 0px, clip]{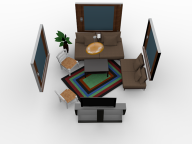}
\caption{Visual comparisons between synthesized scenes using different generators. Left: training a fully connected generator network; Middle: training a sparsely connected generator network; Right: training a combination of sparsely connected generator network and an image-based discriminator.}
\label{Figure:Object:Arrangement:Generator}
\end{figure}

At the heart of our design are
\begin{itemize}
\item the encoder network $\set{G}_{\theta_{E}}: \R^{(d+9)\times n_o}\rightarrow \R^k$, and
\item the decoder network $\set{G}_{\theta}:\R^k \rightarrow \R^{(d+9)\times n_o}$. 
\end{itemize}
Since our matrix encoding of 3D scenes is essentially a vectorized representation (in contrast to an image-based representation), it is natural to use fully connected (FC)-type layers for both the generator network and the encoder network. However, we observed that the naive approach of connecting all pairs of nodes between consecutive layers does not work. Our experiments indicated that this approach easily overfits the training data, so that the generated scenes are of poor quality (See Figure~\ref{Figure:Object:Arrangement:Generator}(a)).


To address this overfitting issue, we propose to use sparsely connected layers. Each node of one layer is only connected to $h$ nodes of the previous layers. In our implementation, we set $h=4$ and randomize the connections, i.e., each node independently connects with a node in the previous layer with probability $h/L$, where $L$ is the number of nodes in the previous layer. As illustrated in Figure\ref{Figure:Network:Architecture}, our network interleaves between sparsely connected layers and fully connected layers. We still keep some fully connected layers to give the network sufficient expressiveness for network fitting. Note that the connections between nodes remain fixed during the training process. 

There are two motivations for using sparsely-connected layers for $\set{G}_{\theta}$. First, patterns in 3D scenes usually involve small groups of objects~\cite{Fisher:2010:CSM,Fisher:2011:CSR,Fisher:2012:ESO}, e.g. chairs and tables, or nightstands and beds, so that sparse relationships between object classes are expected. 
Second, from the perspective of optimization, sparsely-connected networks have significantly reduced model size, and thus tend to avoid overfitting and have improved generalization. In the broader picture, neural networks exhibit exponential expressiveness (c.f.~\cite{NIPS2016_6322}), and training compressed networks yield comparable and sometimes better generalization performance~\cite{DBLP:journals/corr/HanMD15,DBLP:journals/corr/HanPNMTECTD16}. 

Following DCGAN~\cite{DBLP:journals/corr/RadfordMC15}, we set the architecture of $\set{G}_{\theta}$ to the reverse of that of $\set{G}_{\theta_{E}}$. We use VAE-GAN~\cite{Larsen:2016:ABP} for training both the encoder and decoder networks:
\begin{align}
f_{o} & = \frac{1}{N}\sum\limits_{i=1}^{N}\left\|\set{G}_{\theta}\set{G}_{\theta_E}(\overline{M}_i)- \overline{M}_i\right\|_{F}^2 + KL\left(\{\set{G}_{\theta_E}\left(\overline{M}_i\right)\}, p\right)  \nonumber \\
 &\quad+ \lambda \left(\frac{1}{N}\sum\limits_{i=1}^{N}\set{D}_{\phi}\left(\overline{M}_i\right) - E_{\bs{z}\sim p} \set{D}_{\phi}\left[G_{\theta}(\bs{z})\right] \right) .
\label{Eq:Object:Arrangement:Objective}
\end{align}
where the latent distribution $p$ is the standard normal distribution, and the discriminator $\set{D}_{\phi}$ has the same network architecture as $\set{G}_{\theta_{E}}$, except that we replace the latent vector by one node.

\subsection{Image-Based Module}
\label{Subsection:Image:Based:Discriminator}

As discussed in the overview, we introduce a second, image-based discriminator to better capture local arrangement of objects based on geometric detail, such as the spatial relations between chairs and tables and those between nightstands and beds in generated scenes. In our experiments, we found that it is hard to precisely capture such patterns by merely using FC-type discriminators $\set{D}_{\theta}$. As illustrated in Figure~\ref{Figure:Object:Arrangement:Generator}(b), without the image-based module, the learned object arrangement generator $\set{G}_{\theta}$ exhibits various local compatibility issues (e.g., objects intersect with each other). 

Motivated from the fact that CNN-based discriminators can nicely capture local interaction patterns among adjacent objects~\cite{DBLP:journals/corr/RadfordMC15}, we propose to convert the 3D object arrangement into a 2D image by projecting the 3D scene onto the top view, and to then impose a CNN-based discriminator on this 2D image. Specifically, let $\set{P}: \R^{(d+9)\times n_o} \to \R^{r\times r}$ be the projection operator onto an $r\times r$ image ($r = 128$, and details of the projection operator are described in detail below). Denote $\set{D}_{\phi_I}$ as the discriminator for the image-based representation, where $\phi_I$ represents the network parameter. In our experiments, we used ResNet-18~\cite{DBLP:conf/cvpr/HeZRS16}, an established CNN network capable of capturing multi-scale patterns of an image, as our discriminator.

We then use the following discriminator loss for learning the object arrangement generator:
\begin{equation}
f_{I} = \frac{1}{N}\sum\limits_{i=1}^{N} \set{D}_{\phi_I}\left(\set{P}\left[\overline{M}_i\right]\right) - E_{\bs{z}\sim p} \set{D}_{\phi_I}\left(\set{P}\left[\set{G}_{\theta}(\bs{z})\right]\right).
\label{Eq:Image:Based:Objective}
\end{equation}
Rather than projecting the 3D scene to the top view, another option is to convert the scene to a volumetric grid and impose 3D CNN-based discriminator. However, this approach has severe limits on tractable grid resolution (e.g. $64^3$) and cannot accurately resolve local geometric detail. On the other hand, most local correlations are revealed in the top view~\cite{DBLP:journals/corr/abs-1803-08999,Wang:2018:PRM}, which provides sufficient supervision for learning the generator. 

Although it is possible to use a rendering operator for the projection $\set{P}$, as as described by Wang et al.~\cite{Wang:2018:PRM}, we want the image-based discriminator $\set{D}_{\phi_I}$ to provide smooth gradients for the generator $\set{G}_{\theta}$, and such gradients are hard to compute even when using very simple rendering operations. We therefore instead define a fuzzy projection operator $\set{P}$ in terms of summed truncated signed distance fields of objects projected into the top view. Specifically, for each object $o$, let $E_o(M)$ denote the set of points in the plane computed by (1) embedding object $o$ in 3D as encoded by the parameters in $M$, and (2) orthogonally projecting that object onto the $xy$ plane. 
Denote the truncated signed distance function of object $o$ by 
\begin{align*}
d_{o,\delta}:\R^2&\to \R\\
\bs{p} &\mapsto \begin{cases} d[\bs{p}, \partial E_o(M)], & d[\bs{p}, \partial E_o(M)] \leq \delta, \bs{p}\not\in E_o(M)\\
-d[\bs{p}, \partial E_o(M)], & d[\bs{p}, \partial E_o(M)] \leq \delta, \bs{p}\in E_o(M)\\
0, & d[\bs{p}, \partial E_o(M)] > \delta.\end{cases}
\end{align*}
Let $d_{o,\delta}^{I}\in\R^{r\times r}$ be the rasterization of $d_{o,\delta}$ onto an $r\times r$ image. We then define the projection operator as
\begin{equation}
\set{P}(M) := \sum_{o \in M} c_{o}d_{o,M}^{I},
\label{Eq:Projection:Operator}
\end{equation}
where $c_o$ is a class-specific constant associated with object $o$. 
In our implementation, we simply use the index of the category label of $o$ (See Appendix~\ref{Section:Statistics:SUNCG}). For fixed $M$, the gradient of $\set{P}(M)$ with respect to $M$ can be computed; see Appendix~\ref{Section:Projection:Operator} for details.

\subsection{Joint Scene Alignment}
\label{Subsection:Scene:Alignment}

As a preprocessing step, we align all input training scenes, by assigning each scene a rigid transformation and set of permutations, as described in section~\ref{Subsection:Scene:Representation}. We follow the common two-step procedure for establishing consistent transformations (maps) across a collection of objects~\cite{Huber_2002_4190,Huang:2006:RFO,Kim:2012:ECM,Huang:2013:CSM}, namely, we first perform pairwise matching, and then aggregate these pairwise matches into a consistent global alignment of all scenes. A common feature of such two-step approaches is that the second step can effectively remove noisy pairwise matches computed in the first step~\cite{Huang:2013:CSM}, leading to high-quality alignments. In our case, simultaneously optimizing for each scene's optimal rigid transformation and permutations is intractable for large-scale data (i.e. several thousands of scenes). We therefore propose to align the input scenes in a sequential manner, by first optimizing rotations, then translations and finally permutations. 

\paragraph{Pairwise matching.}
Given a pair of scenes $M^i$ and $M^j$, we solve the following optimization problem to determine the optimal transformation $\set{T}_{ij}^{\init} = \left(R_{ij}^{\init}, \bs{t}_{ij}^{\init}\right)$ aligning $M_i$ to $M_j$, as well permutations $\set{S}_{ij}^{\init}$ mapping objects of each class in $M^i$ to their closest match in $M^j$:
\begin{equation}
\set{T}_{ij}^{\init}, \set{S}_{ij}^{\init} = \underset{\set{T}, \set{S}}{\textup{argmin}}\ \left\|\left(\set{T}\circ \set{S}\right)\left(M^i\right)-M^j\right\|_{2,1},
\label{Eq:Pairwise:Alignment:Objective}
\end{equation}
where $\|A\|_{2,1} = \sum\limits_{j=1}^{m}\|\bs{a}_j\|_{\star}, A:= (\bs{a}_1,\cdots, \bs{a}_m)$ is a robust norm used to handle continuous and discrete variations between $M_i$ and $M_j$. 

We solve equation~\eqref{Eq:Pairwise:Alignment:Objective} by combining the method of reweighted least squares~\cite{Daubechies_iterativelyreweighted} and alternating minimization. Since this step is not the major contribution of the paper, we defer the technical details, as well as the precise definition of $\|\|_{\star}$, to Appendix~\ref{Eq:Pairwise:Scene:Alignment}.  

Computing pairwise alignments of all pairs of scenes is infeasible. We follow the procedure of Heath et al.~\shortcite{conf/cvpr/HeathGOAG10} by connecting each scene with $k=64$ neighboring scenes in the training set. To compute these nearest neighbors, we assign to each scene an $n_c$-dimensional vector that counts how many objects of each class appear in the scene, and compute nearest neighbors via L2-distance between these vectors. 

\begin{figure}
\centering
\begin{overpic}[width=1.0\columnwidth]{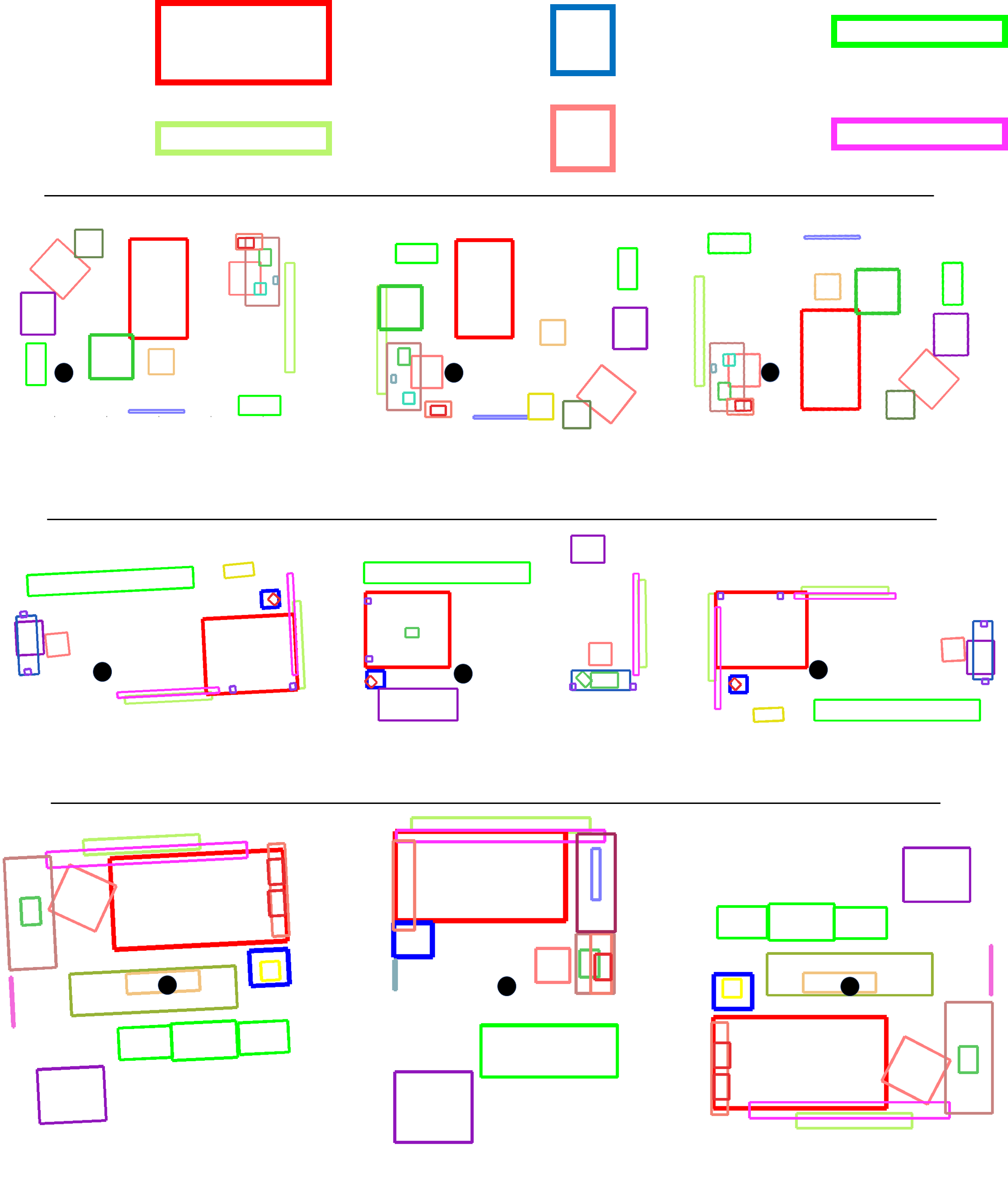}
\put(15,280){Bed:}
\put(86,280){Nightstand:}
\put(165,280){Cabinet:}
\put(-1,255){Window:}
\put(107,255){Chair:}
\put(165,255){Curtain:}
\put(10,175){Source (pairwise)}
\put(112,175){Target}
\put(180,175){Source (joint)}
\put(10,105){Source (pairwise)}
\put(112,105){Target}
\put(180,105){Source (joint)}
\put(10,1){Source (pairwise)}
\put(112,1){Target}
\put(180,1){Source (joint)}
\end{overpic}
\caption{This figure is best visualized in color. Each row presents the alignment of a pair of scans using pairwise alignment (Left) and joint alignment among 5000 bedroom scenes from SUNCG~\protect\cite{song2016ssc} (Right). The black dot indicates the relative translations. For simplicity, we show 2D layouts of each scene from the top view. Different categories of objects possess different colors. We can see that joint alignment, which utilize information from the entire collection to determine the pose of each scene, can rectify pairwise misalignments induced from patterns of uncorrelated object groups. In addition, merely aligning the bed objects is sub-optimal, as the locations of bed exhibit significant variations (See the third row).}
\label{Figure:Joint:Scene:Alignment}
\end{figure}
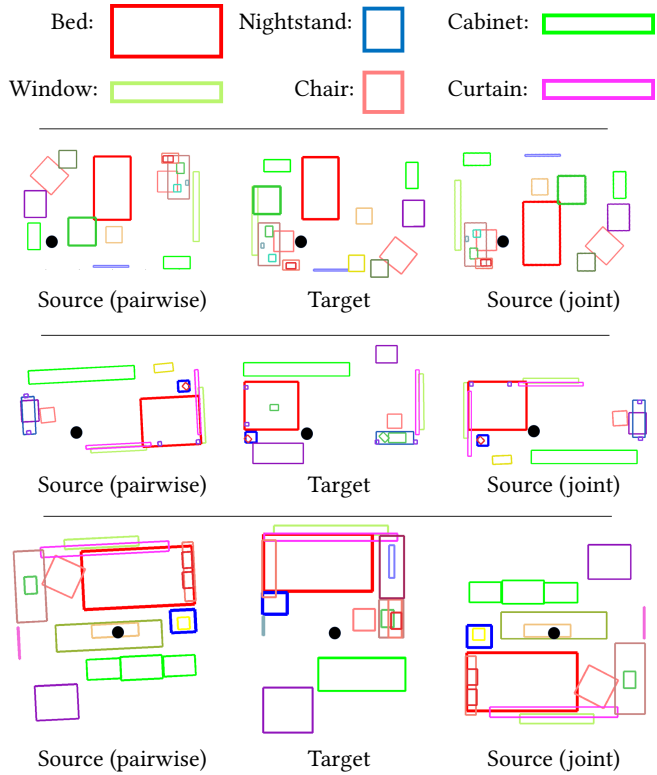

We expect some pairwise alignments to be noisy (see for instance  Figure~\ref{Figure:Joint:Scene:Alignment}). We address this issue by a second, global alignment refinement step (map synchronization). 

\paragraph{Consistent scene alignment.}
We employ state-of-the-art map synchronization techniques for joint optimization of orientations, translations, and permutations associated with the input scenes. For efficiency, we optimize orientations, translations, and permutations in a sequential manner. For rotation synchronization, we employ the method of Chatterjee and Govindu~\shortcite{DBLP:conf/iccv/ChatterjeeG13}, which optimizes the orientation $R_i$ associated with each scene $S_i$ via Huber loss. For translation synchronization, we use the method of Huang et al.~\shortcite{NIPS2017_6744}, which applies truncated least squares to optimize the translation of $\bs{t}_i$ associated with each scene $S_i$. Finally, we employ normalized spectral permutation synchronization~\cite{NIPS2016_6128} to optimize the permutation $\sigma_{i,k}$ associated with each category $c_k$ of each scene $S_i$. Since our approach directly applies these techniques, we refer to these respective papers for technical details.

Note that all of these approaches can tolerate a significant amount of noise in the pairwise alignments. As a result, alignments substantially improve after the map synchronization step (see Figure~\ref{Figure:Joint:Scene:Alignment}). 

\subsection{Network Training}
\label{Subsection:Network:Training}

Finally, given the consistently aligned scenes, we proceed to learn the object arrangement generator $G_{\theta}$, the object arrangement encoder $G_{\theta_{E}}$, and the two discriminators $\set{D}_{\phi}$ and $\set{D}_{\phi_{I}}$. 
Combining equations~\eqref{Eq:Scene:Distance}), \eqref{Eq:Object:Arrangement:Objective}, and \eqref{Eq:Image:Based:Objective}, we arrive at the following objective function:
\begin{align}
\max\limits_{\phi,\phi_{I}} & \min\limits_{\theta, \theta_{E}}\ \frac{1}{N}\sum\limits_{i=1}^{N}\left \|\set{G}_{\theta}\set{G}_{\theta_{E}}\left(\overline{M}_i\right) - \overline{M}_i\right\|_{F}^2 \nonumber \\
&+ \frac{\gamma}{N} \sum_{i=1}^N   \min\limits_{\set{T}_i,\set{S}_i}\left\|\overline{M}_i - 
\left(\set{T}_i \circ \set{S}_i\right)(M_i)\right\|_{F}^2 \nonumber \\
& +\lambda \left(\frac{1}{N}\sum\limits_{i=1}^{N}\set{D}_{\phi}\left(\overline{M}_i\right) - E_{\bs{z}\sim p}\set{D}_{\phi}\left[\set{G}_{\theta}(\bs{z})\right]\right) \nonumber\\
&+ KL\left(\left\{\set{G}_{\theta_E}\left(\overline{M}_i\right)\right\}, p\right) \nonumber \\
& +  \mu\left(\frac{1}{N}\sum\limits_{i=1}^{N}\set{D}_{\phi_I}\left(\set{P}(\overline{M}_i)\right) - E_{\bs{z}\sim p}\set{D}_{\phi_I}\left(\set{P}\left[\set{G}_{\theta}(\bs{z})\right]\right)\right).
\label{Eq:Alter:min}
\end{align}
In this paper, we set $\lambda = 1$, $\mu = 1$, and $\gamma = 100$. The large value in $\gamma$ ensures that $\overline{M}_i$ and $M_i$ encode approximately the same scene. 

Equation~\eqref{Eq:Alter:min} is challenging to solve since the objective function is highly non-convex (even when the discriminators are fixed). We again apply alternating minimization for optimization, so that each step solves an easier optimization sub-problem. 

\subsubsection{Alternating Minimization}

We perform two levels of alternating minimization. The first level alternates between optimizing $\left\{\overline{M}_i, \set{S}_i, \set{T}_i, \theta, \theta_{E}, \theta_I\right\}$ and the discriminators $\left\{\phi, \phi_{I}\right\}$. In the former case, a second level of alternation switches between optimizing $\theta$,$\theta_{E}$, the $\overline{M}_i$, the $\set{S}_i$ and the $\set{T}_i$. 

\paragraph{Generator optimization.}
When $\phi$, $\phi_I$, the $\set{S}_i$, the $\set{T}_i$ and the $\set{M}_i$ are fixed, equation~\eqref{Eq:Alter:min} reduces to 
\begin{align}
\min\limits_{\theta, \theta_{E}} &\  \frac{1}{N}\sum\limits_{i=1}^{N} \left\|\set{G}_{\theta}\set{G}_{\theta_{E}}\left(\overline{M}_i\right) 
- \overline{M}_i\right\|_{F}^2\nonumber\\ 
& -E_{\bs{z}\sim p}\set{D}_{\phi}\left(\set{G}_{\theta}\left[\bs{z})\right]\right)- \mu E_{\bs{z}\sim p}\set{D}_{\phi_I}\left(\set{P}\left[\set{G}_{\theta}(\bs{z})\right]\right).
\label{Eq:Encoder:Generator:Opt}
\end{align}
We apply ADAM~\cite{journals/corr/KingmaB14} for optimization. In all of our experiments, we trained $\theta$ and $\theta_E$ for two epochs and then moved to optimize other variables. 


\paragraph{Latent variable optimization.}
When $\theta$, $\theta_E$, $\phi$, $\phi_I$, the $\set{S}_i$ and the $\set{T}_i$ are fixed, we can optimize $\overline{M}_i$ for each scene in isolation:
\begin{align*}
\min\limits_{\overline{M}_i}\ & \left\|\set{G}_{\theta}\set{G}_{\theta_E}
\left(\overline{M}_i\right)-\overline{M}_i\right\|_F^2 + \gamma\left\|\overline{M}_i-\set{T}_i\left(\set{S}_i(M_i)\right)\right\|_F^2 \\
& + \lambda \set{D}_{\phi}\left(\overline{M}_i\right) + \mu \set{D}_{\phi_I}\left(\set{P}(\overline{M}_i)\right)
\end{align*}
We apply ADAM~\cite{journals/corr/KingmaB14} for optimization. Since the value of $\gamma$ is large, the convex potential $\gamma\left\|\overline{M}_i-\set{S}_i(M_i)\right\|_F^2$ strongly dominates the other terms. In our experiments, we found this step usually converges in 8-12 iterations. 

\paragraph{Permutation optimization.}
When the $\overline{M}_i$, the $\set{T}_i$, $\theta$, $\theta_E$, $\phi$ and $\phi_I$ are fixed, we can optimize $\sigma_{i,k}$ in each $\set{S}_i$ in isolation:
\begin{equation}
\sigma_{i,k}^{\star} = \underset{\sigma_{i,k}\in S_{m_k}}{\textup{argmin}} \left\|\set{T}_i^{-1}(\overline{M}_{i,k}) - M_{i,k} \sigma_{i,k}\right\|_F^2,
\label{Eq:Perm:Opt}
\end{equation}
for $1\leq k \leq n_c, 1\leq i \leq N.$ It is easy to see that equation~\eqref{Eq:Perm:Opt} is equivalent to
$$
\sigma_{i,k}^{\star} = \underset{\sigma_{i,k}\in S_{m_k}}{\textup{argmax}}\left\langle \sigma_{i,k}, \set{T}_i^{-1}\left(\overline{M}_{i,k}\right)M_{i,k}^{T} \right\rangle 
$$
which is a linear assignment problem, and can be solved exactly using the Hungarian algorithm.

\paragraph{Transformation optimization.}
When the $\overline{M}_i$, the $\set{S}_i$, $\theta$, $\theta_E$, $\phi$ and $\phi_I$ are fixed, we can optimize each $\set{T}_i$ in isolation:
\begin{equation}
\set{T}_i^{\star} = \underset{\set{T}_i}{\textup{argmin}}\ \left\|\overline{M}_i - \set{T}_i\left(\set{S}_i(M_i)\right)\right\|_{F}^2.   
\label{Eq:Trans:Opt}
\end{equation}
Equation~\eqref{Eq:Trans:Opt} can be formulated as rigid point cloud alignment with known correspondences (orthogonal Procrustes), and we use the closed-form solution described by Horn~\shortcite{Horn87closed-formsolution}.

\paragraph{Discriminator optimization.}
Finally, when the $\overline{M}_i$, the $S_i$, $\theta$ and $\theta_{E}$ are fixed, the discriminators can be optimized independently as follows:
\begin{align}
\min\limits_{\phi}&\ \frac{1}{N}\sum\limits_{i=1}^{N} D_{\phi}\left(\overline{M}_i\right) - E_{\bs{z}\sim p}\set{D}_{\phi}\left(G_{\theta}(\bs{z})\right) \nonumber \\
\min\limits_{\phi_I}&\ \frac{1}{N}\sum\limits_{i=1}^{N}\set{D}_{\phi_{I}}\left[\set{P}\left(\overline{M}_i\right)\right] - E_{\bs{z}\sim p}\set{D}_{\phi_I}\left(\set{P}\left[\set{G}_{\theta}(\bs{z})\right]\right).\nonumber 
\end{align}
In our experiments, we trained both discriminators for 10 epochs in each alternating minimization.

\paragraph{Termination of alternating minimization.}
In all of our experiments, we use $t_{max}^{\textup{inner}} = 10$ iterations for the inner alternating minimization (i.e., of $\theta$,$\theta_{E}$, the $\overline{M}_i$, the $\set{S}_i$, and the $\set{T}_i$) and $t_{\max}^{\textup{outer}} = 10$ iterations of the outer alternating minimization (between the set of preceding variables and $\{\phi, \phi_{I}\}$). 
\section{Experimental Evaluation}
\label{Section:Experimental:Evaluation}

In this section, we present an experimental evaluation of the proposed approach. In Section~\ref{Section:Experimental:Setup}, we describe the experimental setup. From Section~\ref{Section:Experimental:Results} to Section~\ref{Section:Ablation:Study}, we analyze the results of our approach. Section~\ref{Section:Scene:Interpolation} and Section~\ref{Section:Scene:Completion} present the applications our approach in scene interpolation and scene completion, respectively.

\subsection{Experimental Setup}
\label{Section:Experimental:Setup}


\begin{figure*}
\centering
\includegraphics[width=0.192\textwidth, trim=15px 15px 15px 0px, clip]{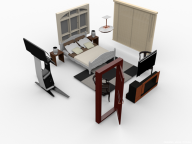}
\includegraphics[width=0.192\textwidth, trim=15px 15px 15px 0px, clip]{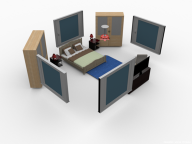}
\includegraphics[width=0.192\textwidth, trim=15px 15px 15px 0px, clip]{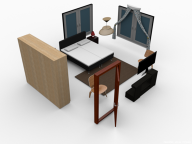}
\includegraphics[width=0.192\textwidth, trim=15px 15px 15px 0px, clip]{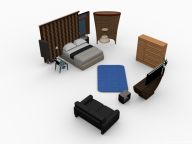}
\includegraphics[width=0.192\textwidth, trim=15px 15px 15px 0px, clip]{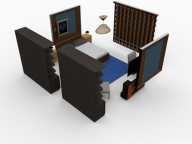}
\includegraphics[width=0.192\textwidth, trim=15px 15px 15px 0px, clip]{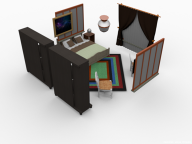}
\includegraphics[width=0.192\textwidth, trim=15px 15px 15px 0px, clip]{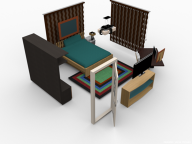}
\includegraphics[width=0.192\textwidth, trim=15px 15px 15px 0px, clip]{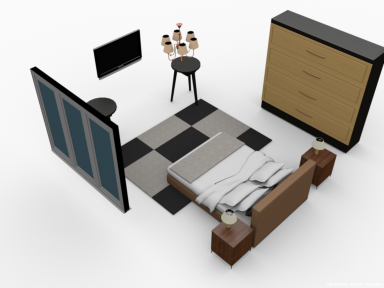}
\includegraphics[width=0.192\textwidth, trim=15px 15px 15px 0px, clip]{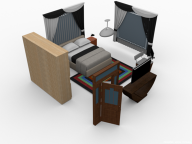}
\includegraphics[width=0.192\textwidth, trim=15px 15px 15px 0px, clip]{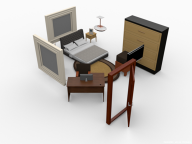}
\includegraphics[width=0.192\textwidth, trim=15px 15px 15px 0px, clip]{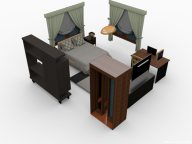}
\includegraphics[width=0.192\textwidth, trim=15px 15px 15px 0px, clip]{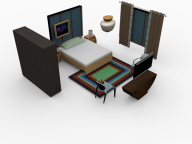}
\includegraphics[width=0.192\textwidth, trim=15px 15px 15px 0px, clip]{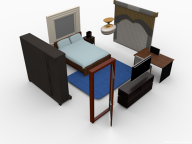}
\includegraphics[width=0.192\textwidth, trim=15px 15px 15px 0px, clip]{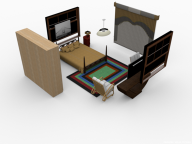}
\includegraphics[width=0.192\textwidth, trim=15px 15px 15px 0px, clip]{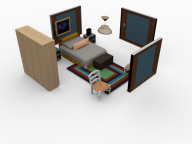}
\includegraphics[width=0.192\textwidth, trim=15px 30px 15px 0px, clip]{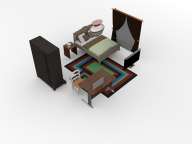}
\includegraphics[width=0.192\textwidth, trim=15px 30px 15px 0px, clip]{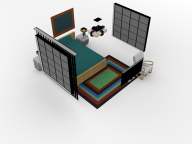}
\includegraphics[width=0.192\textwidth, trim=15px 30px 15px 0px, clip]{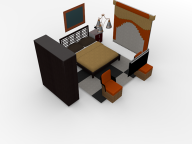}
\includegraphics[width=0.192\textwidth, trim=15px 30px 15px 0px, clip]{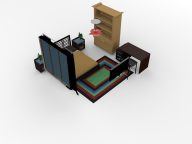}
\includegraphics[width=0.192\textwidth, trim=15px 30px 15px 0px, clip]{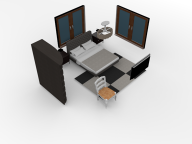}
\includegraphics[width=0.192\textwidth, trim=15px 30px 15px 0px, clip]{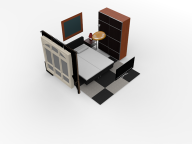}
\includegraphics[width=0.192\textwidth, trim=15px 30px 15px 0px, clip]{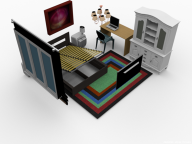}
\includegraphics[width=0.192\textwidth, trim=15px 30px 15px 0px, clip]{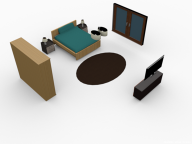}
\includegraphics[width=0.192\textwidth, trim=15px 30px 15px 0px, clip]{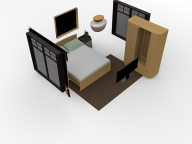}
\includegraphics[width=0.192\textwidth, trim=15px 30px 15px 0px, clip]{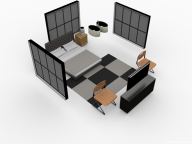}
\includegraphics[width=0.192\textwidth, trim=15px 30px 15px 0px, clip]{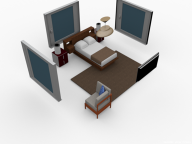}
\includegraphics[width=0.192\textwidth, trim=15px 30px 15px 0px, clip]{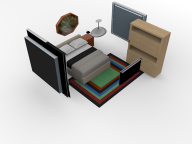}
\includegraphics[width=0.192\textwidth, trim=15px 30px 15px 0px, clip]{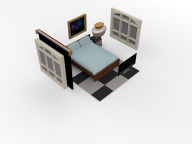}
\includegraphics[width=0.192\textwidth, trim=15px 30px 15px 0px, clip]{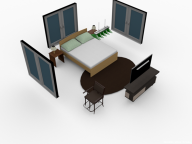}
\includegraphics[width=0.192\textwidth, trim=15px 30px 15px 0px, clip]{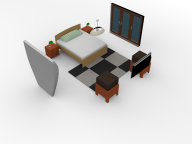}
\caption{Randomly generated scenes of bedrooms.}
\label{Figure:Bedroom:Synthesis}
\end{figure*}

\begin{figure*}
\centering
\centering
\includegraphics[width=0.192\textwidth, trim=25px 35px 25px 0px, clip]{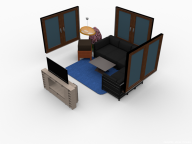}
\includegraphics[width=0.192\textwidth, trim=25px 35px 25px 0px, clip]{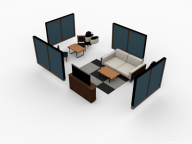}
\includegraphics[width=0.192\textwidth, trim=25px 35px 25px 0px, clip]{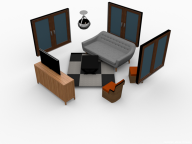}
\includegraphics[width=0.192\textwidth, trim=25px 35px 25px 0px, clip]{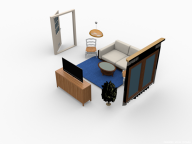}
\includegraphics[width=0.192\textwidth, trim=25px 35px 25px 0px, clip]{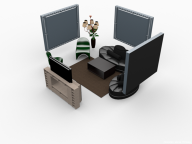}
\includegraphics[width=0.192\textwidth, trim=40px 35px 20px 0px, clip]{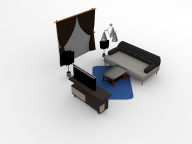}
\includegraphics[width=0.192\textwidth, trim=25px 35px 25px 0px, clip]{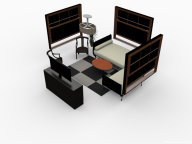}
\includegraphics[width=0.192\textwidth, trim=25px 35px 25px 0px, clip]{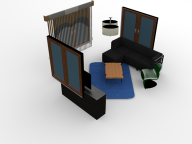}
\includegraphics[width=0.192\textwidth, trim=25px 35px 25px 0px, clip]{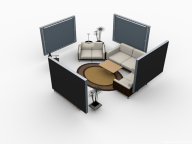}
\includegraphics[width=0.192\textwidth, trim=25px 35px 25px 0px, clip]{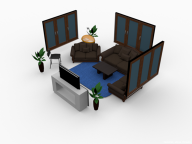}
\includegraphics[width=0.192\textwidth, trim=25px 30px 20px 0px, clip]{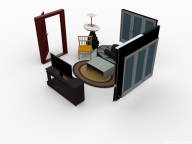}
\includegraphics[width=0.192\textwidth, trim=25px 30px 20px 0px, clip]{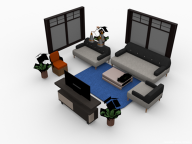}
\includegraphics[width=0.192\textwidth, trim=25px 30px 20px 0px, clip]{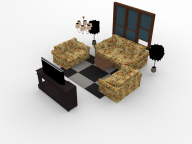}
\includegraphics[width=0.192\textwidth, trim=25px 30px 20px 0px, clip]{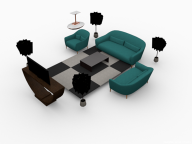}
\includegraphics[width=0.192\textwidth, trim=25px 30px 20px 0px, clip]{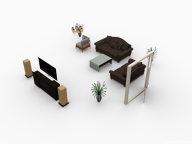}
\includegraphics[width=0.192\textwidth, trim=25px 30px 25px 0px, clip]{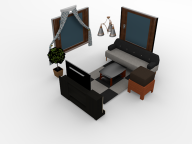}
\includegraphics[width=0.192\textwidth, trim=25px 30px 25px 0px, clip]{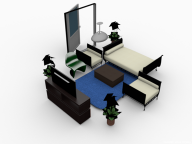}
\includegraphics[width=0.192\textwidth, trim=25px 30px 25px 0px, clip]{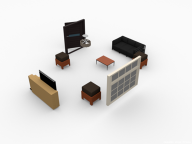}
\includegraphics[width=0.192\textwidth, trim=25px 30px 25px 0px, clip]{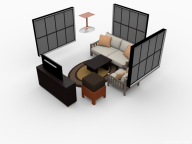}
\includegraphics[width=0.192\textwidth, trim=25px 30px 25px 0px, clip]{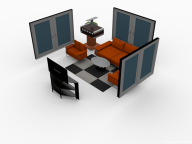}
\includegraphics[width=0.192\textwidth, trim=25px 30px 25px 0px, clip]{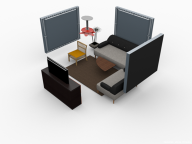}
\includegraphics[width=0.192\textwidth, trim=25px 30px 25px 0px, clip]{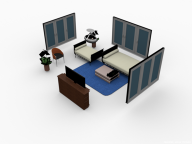}
\includegraphics[width=0.192\textwidth, trim=25px 30px 25px 0px, clip]{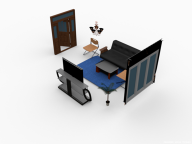}
\includegraphics[width=0.192\textwidth, trim=25px 30px 25px 0px, clip]{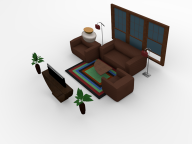}
\includegraphics[width=0.192\textwidth, trim=25px 30px 25px 0px, clip]{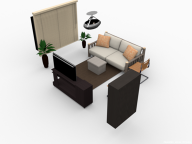}
\includegraphics[width=0.192\textwidth, trim=25px 30px 00px 0px, clip]{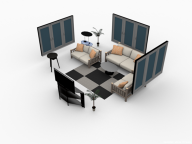}
\includegraphics[width=0.192\textwidth, trim=25px 30px 00px 0px, clip]{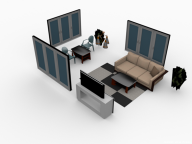}
\includegraphics[width=0.192\textwidth, trim=25px 30px 00px 0px, clip]{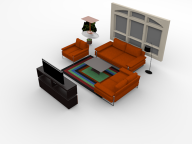}
\includegraphics[width=0.192\textwidth, trim=25px 30px 00px 0px, clip]{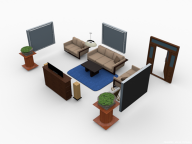}
\includegraphics[width=0.192\textwidth, trim=25px 30px 00px 0px, clip]{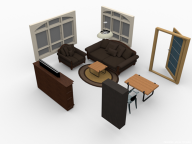}
\caption{Randomly generated scenes of living rooms.}
\label{Figure:Livingroom:Synthesis}
\end{figure*}
\paragraph{Dataset}
We perform experimental evaluation on two datasets extracted from SUNCG~\cite{song2016ssc}: \emph{Bedroom} and \emph{Living Room}. SUNCG is a large database of virtual 3D scenes created by users of the online Planner5d interior design tool~\cite{Planner5d}. 
The dataset contains over 45,000 3D scenes, with each scene segmented into individual rooms labeled with a room type. In this work, we used the $30$ most frequent classes. Please refer to Appendix~\ref{Section:Statistics:SUNCG} for the names of these classes and other statistics. For each class, we constrain that the maximum repetition of an object category is $4$. We use this criteria to gather all suitable scenes (i.e., all objects fall in those $30$ most frequent classes and the largest repetition count is $4$). In total, we collect $5922$ and $5688$ Bedroom and Living Room scenes, respectively. For both datasets, we randomly select $5000$ scenes for training. The remaining scenes are left for testing. We directly use the 3D models associated with SUNCG as the shape database for our approach.

\paragraph{Baseline approaches}
Since we are unaware of any existing methods that have the exact input and output settings as our approach, we perform evaluation against variants of our approach:
\begin{itemize}
\item \textsl{Baseline I:} The first baseline removes the image-based module and only applies VAE-GAN on the object arrangement representation.
\item \textsl{Baseline II:} The second baseline further modifies Baseline I by replacing sparsely connected layers with fully connected layers. 
\end{itemize}

In Section~\ref{Section:Scene:Completion}, we compare our approach with two state-of-the-art data-driven scene synthesis~\cite{Fisher:2012:ESO,Kermani:2016:LSS} for the task of scene completion.

\subsection{Experimental Results}
\label{Section:Experimental:Results}

Figure~\ref{Figure:Best:Figure}, Figure~\ref{Figure:Bedroom:Synthesis}, and Figure~\ref{Figure:Livingroom:Synthesis} show randomly generated scenes using our approach. The overall quality is high and matches that of the training data (the quantitative evaluation is presented in Section~\ref{Section:Perceptual:Study}). The synthesized scenes are also diverse, exhibiting large variations in number of objects in a scene, spatial layouts, and correlated groups of objects. 

Figure~\ref{Figure:Best:Figure} compares the generated scenes with their closest scenes in the training data. Here we simply employ the Euclidean distance in the latent scene space for computing closest scenes. We can see that the generated scenes exhibit noticeable variations in spatial object layout and object existence. This means that our approach learns meaningful patterns in the training data and uses them for scene synthesis, instead of merely memorizing the data.

\subsection{Perceptual Study}
\label{Section:Perceptual:Study}

\begin{figure}
\centering
\begin{overpic}[width=1.0\columnwidth]{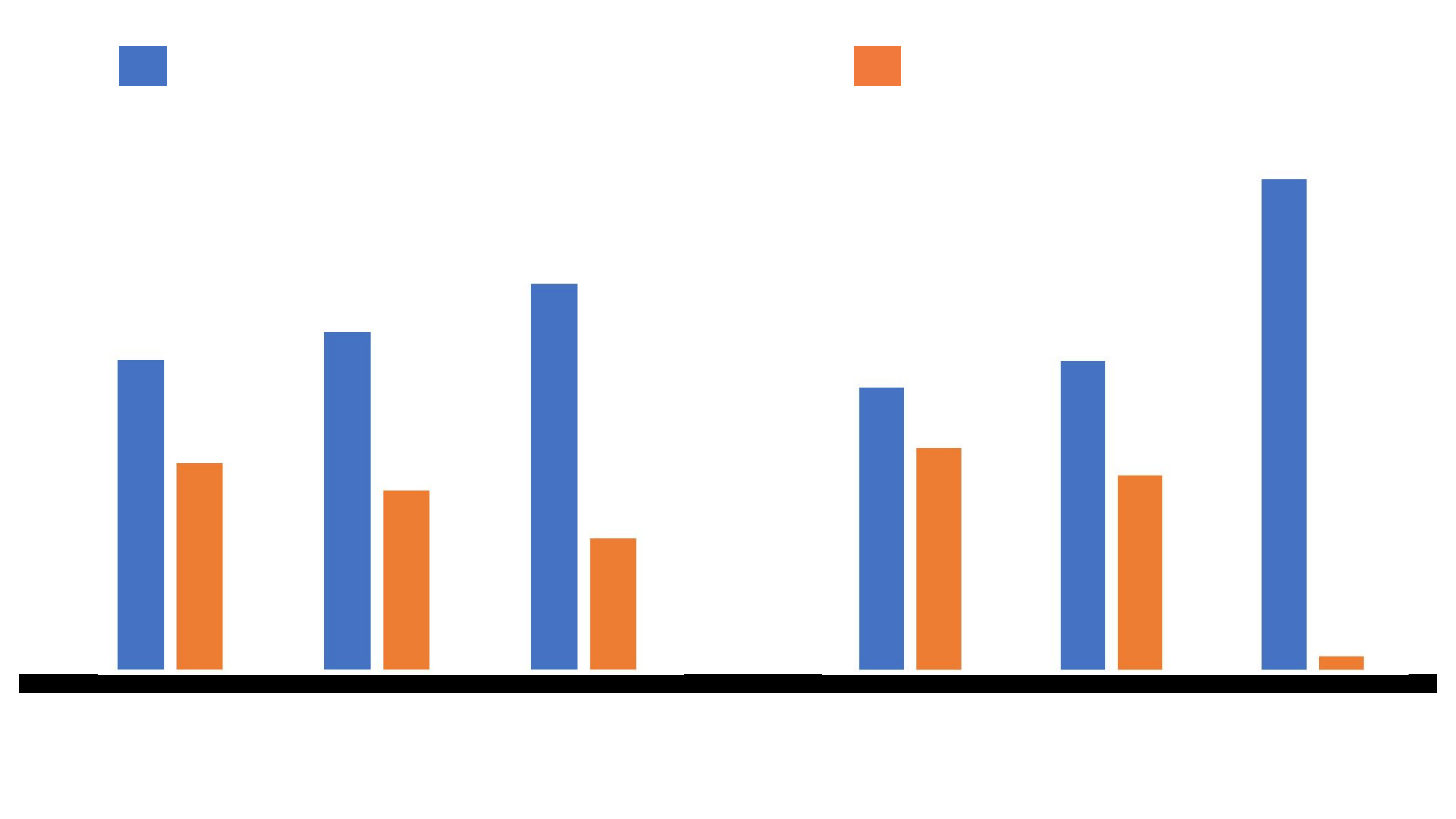}
\put(40,-8){Bedroom}
\put(151,-8){Livingroom Room}
\put(20,5){\footnotesize{Ours}}
\put(47,5){\footnotesize{BaselineI}}
\put(83,5){\footnotesize{BaselineII}}
\put(143,5){\footnotesize{Ours}}
\put(169,5){\footnotesize{BaselineI}}
\put(203,5){\footnotesize{BaselineII}}
\put(14,82){\footnotesize{60.00\%}}
\put(28,63){\footnotesize{40.00\%}}
\put(50,86){\footnotesize{65.33\%}}
\put(64,57){\footnotesize{34.67\%}}
\put(86,93){\footnotesize{74.67\%}}
\put(99,50){\footnotesize{25.33\%}}
\put(136,74){\footnotesize{56.00\%}}
\put(152,63){\footnotesize{44.00\%}}
\put(172,77){\footnotesize{61.33\%}}
\put(186,58){\footnotesize{38.67\%}}
\put(208,108){\footnotesize{97.33\%}}
\put(221,30){\footnotesize{2.67\%}}
\put(30,123){\small{: Real/Real and Syn./Syn.}}
\put(155,123){\small{: Real/Syn. and Syn./Real}}
\put(0,13){\small{Variance:}}
\put(31,13){\footnotesize{$\pm$5.96\%}}
\put(66,13){\footnotesize{$\pm$2.67\%}}
\put(100,13){\footnotesize{$\pm$18.09\%}}
\put(155,13){\footnotesize{$\pm$9.98\%}}
\put(188,13){\footnotesize{$\pm$2.67\%}}
\put(221,13){\footnotesize{$\pm$3.26\%}}
\end{overpic}
\caption{User study on generated scenes using our approach and two baseline approaches on the Bedroom and Living Room datasets. Blue bar indicates the percentage of Real/Synthetic pairs that are marked as Real/Synthetic. Likewise, Orange bar indicates the percentage of Real/Synthetic pairs that are marked as Synthetic/Real. }
\label{Figure:Percenptual:Study}
\end{figure}
We have performed a user study to evaluate the visual quality of our approach versus baseline approaches by following the protocol described in~\cite{DBLP:journals/corr/ShrivastavaPTSW16}. Specifically, for each approach and each dataset we generate 20 scenes. For each scene, we extract the closest scene in the training data. We then present these 20 pairs to users and ask them to choose the scene which they think is generated. We averaged the results over 5 users in the age of 20-50 with minimal background in Computer Graphics. Each study is summarized using a pair of percentages $(a,100-a)$, where $a$ indicates the percentage that scenes in the synthetic data are marked as generated. 

Figure~\ref{Figure:Percenptual:Study} plots the statistics among our approach and the two baseline approaches. We can see that for both the Bedroom and Living Room datasets, our approach yields significantly better results than baseline approaches. In other words, the design choices of using sparsely connected layers and image-based discriminator loss are important for learning to generate realistic scenes. In addition, our approach achieved $60\%\slash 40\%$ and $56\%\slash 44\%$ on Bedroom and Living Room, respectively. Given that the training data mostly consists of high-quality user designed scenes, these numbers are quite encouraging, as more than $30\%$ of the time, users favored our synthesis results rather than user designed scenes. The study was conducted on Amazon Mechanical Turk, and for each approach, we showed 15 different comparisons in the survey and collected 20 surveys from different users. 

\subsection{What are Learned}
\label{Section:What:are:Learned}

In this section, we analyze the performance of our approach by studying what are learned by the neural network. Our protocol is to assess whether important distributions about objects and object pairs in the training data are learned by the network, i.e., if the generated scenes have a distribution similar to the training data. 

\paragraph{Absolute locations of objects} We first evaluate whether our approach learns important distributions of the absolute locations of the objects. To this end, we have tested the distributions of absolute locations of Nightstand, Bed, Window and Television for the Bedroom dataset, and Door, Window, Rug and Plant for the Living Room dataset. For each object, we calculate the distributions in the training data (with respect to the aligned scenes $\overline{M}_i$) versus  5000 randomly generated scenes. 
For simplicity, we only plot the marginal distribution on the x-y plane (or the top view), which captures most of the signals. 

\begin{figure}
\begin{tabular}{c|c|c|c}
\multicolumn{4}{c}{Bedroom. \ \ Top:Training, \ Middle: No-align,\  Bottom: Ours.} \\
Nightstand & Bed & Window & Television \\
\includegraphics[width=0.10\textwidth, trim=30px 30px 30px 30px, clip]{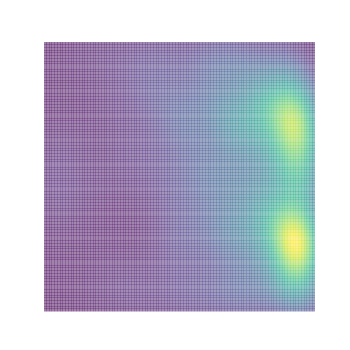}
&
\includegraphics[width=0.10\textwidth, trim=30px 30px 30px 30px, clip]{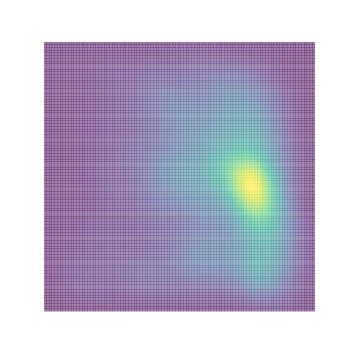}
&
\includegraphics[width=0.10\textwidth, trim=30px 30px 30px 30px, clip]{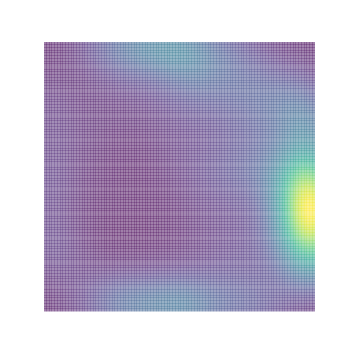}
&
\includegraphics[width=0.10\textwidth, trim=30px 30px 30px 30px, clip]{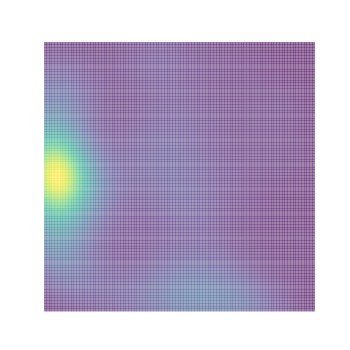}
\\
\includegraphics[width=0.10\textwidth, trim=30px 30px 30px 30px, clip]{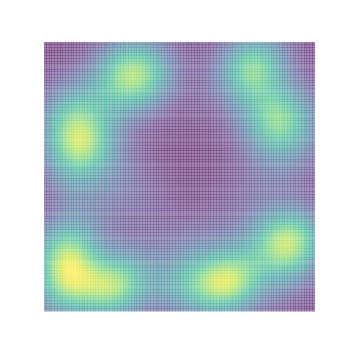}
&
\includegraphics[width=0.10\textwidth, trim=30px 30px 30px 30px, clip]{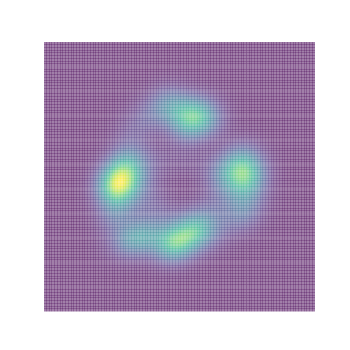}
&
\includegraphics[width=0.10\textwidth, trim=30px 30px 30px 30px, clip]{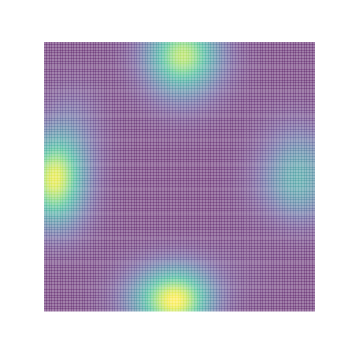}
&
\includegraphics[width=0.10\textwidth, trim=30px 30px 30px 30px, clip]{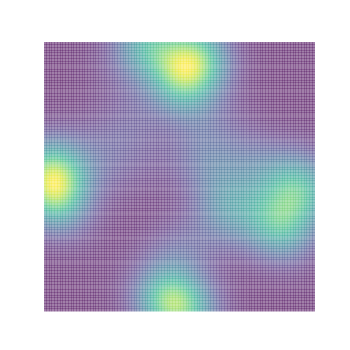}
\\
\includegraphics[width=0.10\textwidth, trim=30px 30px 30px 30px, clip]{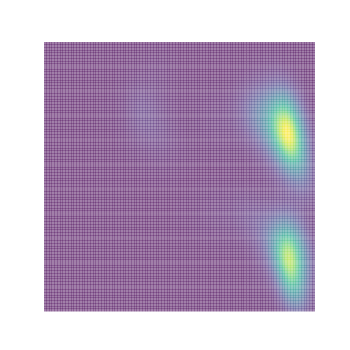}
& 
\includegraphics[width=0.10\textwidth, trim=30px 30px 30px 30px, clip]{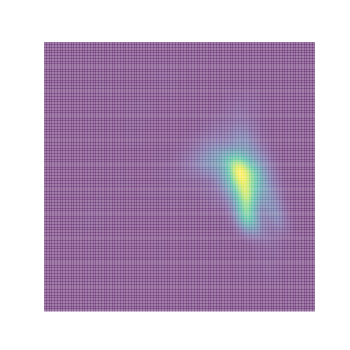}
&
\includegraphics[width=0.10\textwidth, trim=30px 30px 30px 30px, clip]{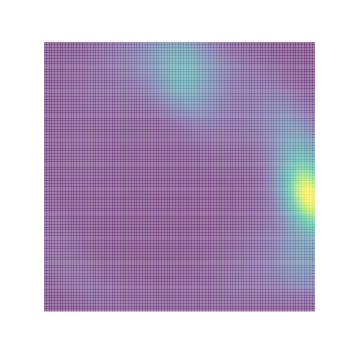}
&
\includegraphics[width=0.10\textwidth, trim=30px 30px 30px 30px, clip]{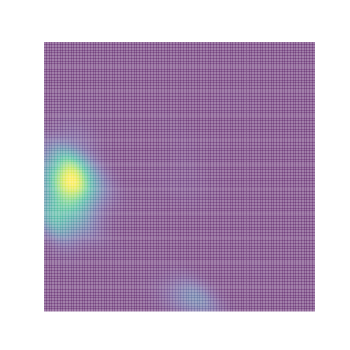}

\\ \hline
\multicolumn{4}{c}{Livingroom. \ \ Top:Training, \ Middle: No-align,\  Bottom: Ours.} \\
Door & Window & Rug & Plant \\
\includegraphics[width=0.10\textwidth, trim=30px 30px 30px 30px, clip]{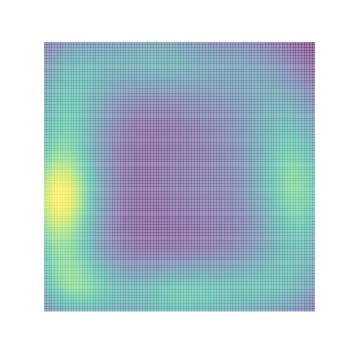}
&
\includegraphics[width=0.10\textwidth, trim=30px 30px 30px 30px, clip]{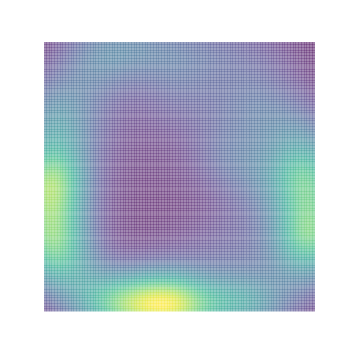}
&
\includegraphics[width=0.10\textwidth, trim=30px 30px 30px 30px, clip]{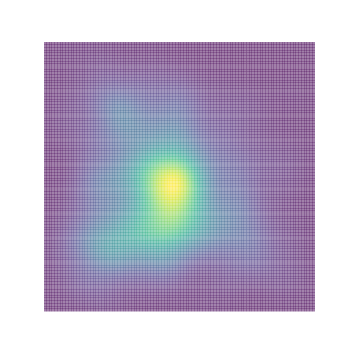}
&
\includegraphics[width=0.10\textwidth, trim=30px 30px 30px 30px, clip]{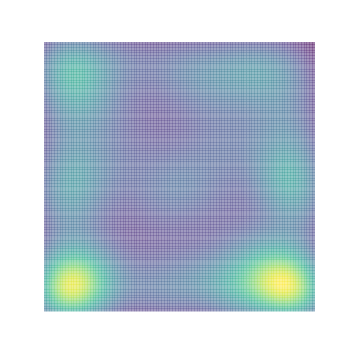}
\\
\includegraphics[width=0.10\textwidth, trim=30px 30px 30px 30px, clip]{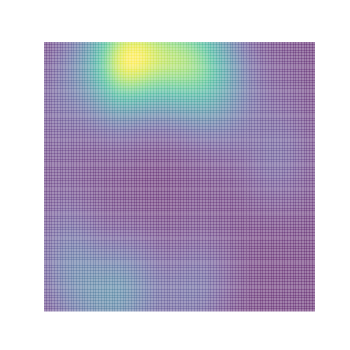}
&
\includegraphics[width=0.10\textwidth, trim=30px 30px 30px 30px, clip]{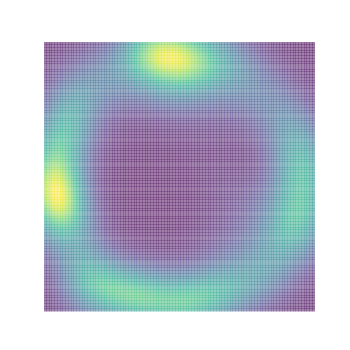}
&
\includegraphics[width=0.10\textwidth, trim=30px 30px 30px 30px, clip]{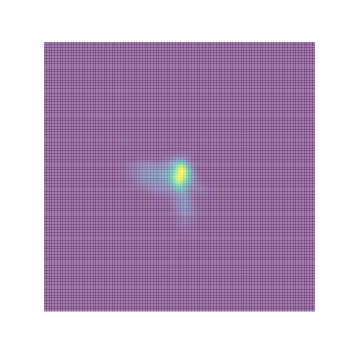}
&
\includegraphics[width=0.10\textwidth, trim=30px 30px 30px 30px, clip]{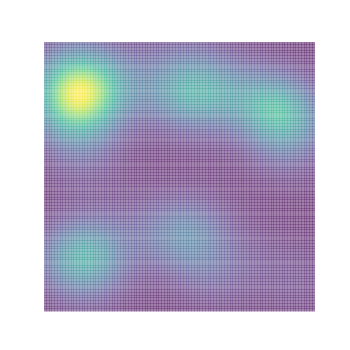}
\\
\includegraphics[width=0.10\textwidth, trim=30px 30px 30px 30px, clip]{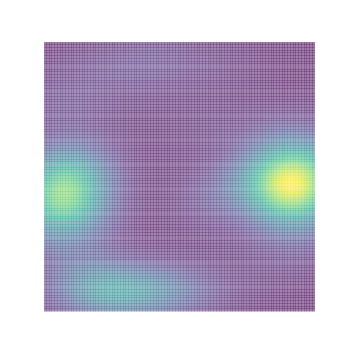}
& 
\includegraphics[width=0.10\textwidth, trim=30px 30px 30px 30px, clip]{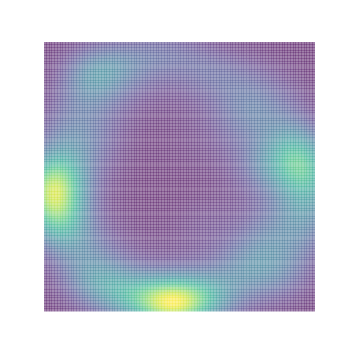}
& 
\includegraphics[width=0.10\textwidth, trim=30px 30px 30px 30px, clip]{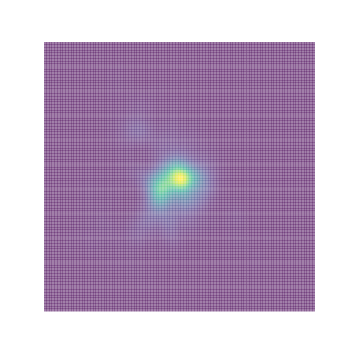}
&
\includegraphics[width=0.10\textwidth, trim=30px 30px 30px 30px, clip]{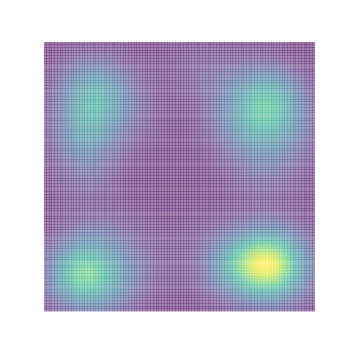}
\\
\end{tabular}
\caption{Distributions of absolute locations of selected classes. Top three rows: Distributions of selected classes in Bedroom for training data, synthesized scenes with no global scene alignment, and our synthesized scenes respectively (from left to right: Nightstand, Bed, Window and Television). Bottom two rows: Top three rows: Distributions of selected classes in Livingroom for training data, synthesized scenes with no global scene alignment, and our synthesized scenes respectively (from left to right: Door, Window, Rug, Plant.)}
\label{Figure:Absolute:Position:Distribution}
\end{figure}

As illustrated in Figure~\ref{Figure:Absolute:Position:Distribution}, the distributions between synthesized scenes of our approach and training scenes are fairly close. In particular, on Window, the difference between the distributions are not easy to identify. The two distributions are less similar on Plant. An explanation is that there are fewer instances of Plant in the training data than other categories, and thus the generalization behavior performs less well. 

\paragraph{Pairwise correlations of objects}

We proceed to evaluate whether important pairwise distributions between objects are learned properly by our generator. Similar to absolute locations of objects, we plot the distributions of the relative location between the second object and the first object. Here the relative location is evaluated with respect to the coordinate system, whose origin is given by the point on the boundary of the bounding box, and whose direction to the bounding box center aligns with the front orientation. We plot the heat-map of the distributions. In this experiment, we consider Desk/Chair, Bed/Nightstand, Bed/Television, and Chair/Computer for Bedroom, and Sofa/Table, Table/Television, Plant/Sofa, and Sofa/Television for Living Room. If there are multiple pairs on one scene, we only extract the pairs with closest spatial distance.
Similar to case of absolute locations of objects, we collect statistics from 5000 training scenes and from 5000 randomly generated scenes. 

\begin{figure}
\begin{tabular}{c|c|c|c}
\multicolumn{4}{c}{Bedroom. \ \ Top:Training, \ Middle: No-align,\  Bottom: Ours.} \\
Desk-Chair & Bed-Night. & Bed-TV& Chair-Computer \\
\includegraphics[width=0.10\textwidth, trim=30px 30px 30px 30px, clip]{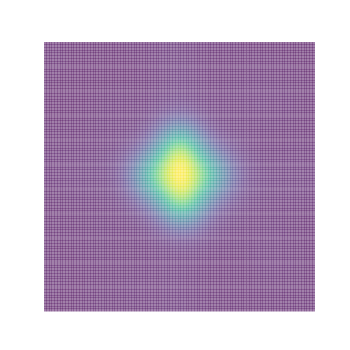}
&
\includegraphics[width=0.10\textwidth, trim=30px 30px 30px 30px, clip]{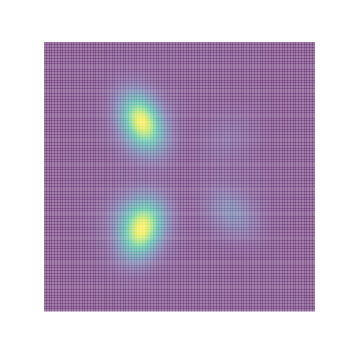}
&
\includegraphics[width=0.10\textwidth, trim=30px 30px 30px 30px, clip]{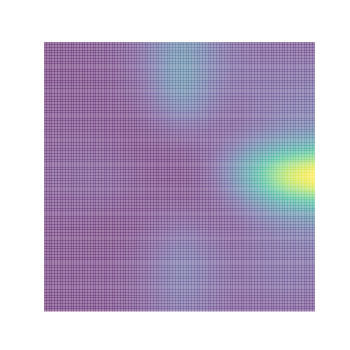}
&
\includegraphics[width=0.10\textwidth, trim=30px 30px 30px 30px, clip]{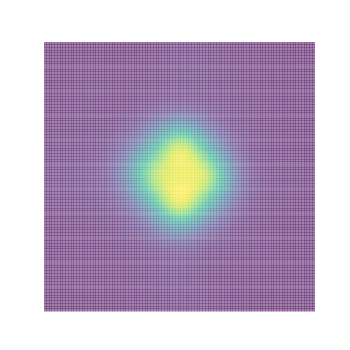}
\\
\includegraphics[width=0.10\textwidth, trim=30px 30px 30px 30px, clip]{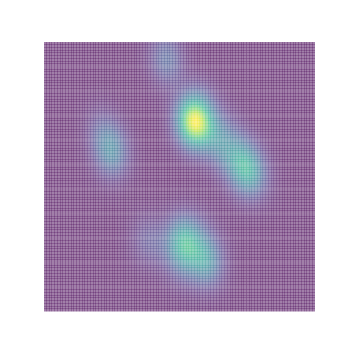}
&
\includegraphics[width=0.10\textwidth, trim=30px 30px 30px 30px, clip]{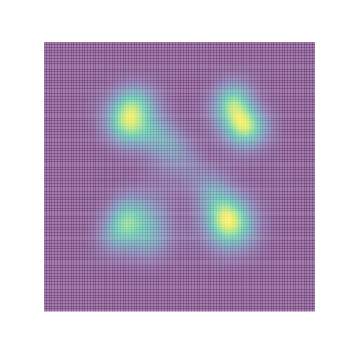}
&
\includegraphics[width=0.10\textwidth, trim=30px 30px 30px 30px, clip]{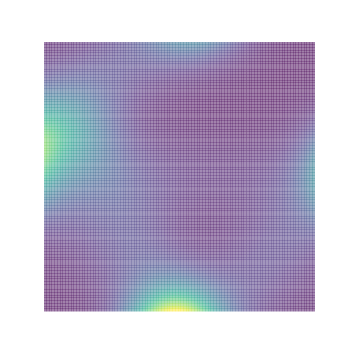}
&
\includegraphics[width=0.10\textwidth, trim=30px 30px 30px 30px, clip]{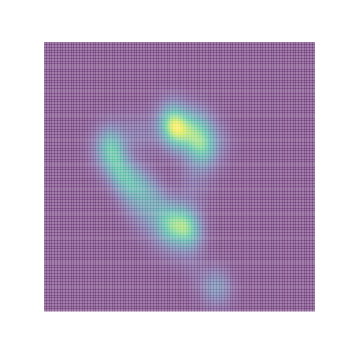}
\\
\includegraphics[width=0.10\textwidth, trim=30px 25px 25px 30px, clip]{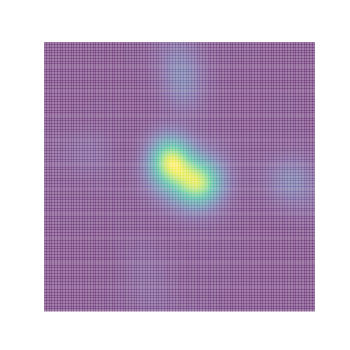}
&
\includegraphics[width=0.10\textwidth, trim=30px 25px 25px 30px, clip]{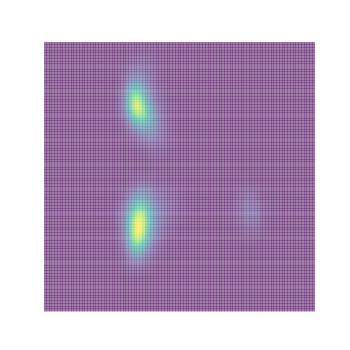}
&
\includegraphics[width=0.10\textwidth, trim=30px 25px 25px 30px, clip] {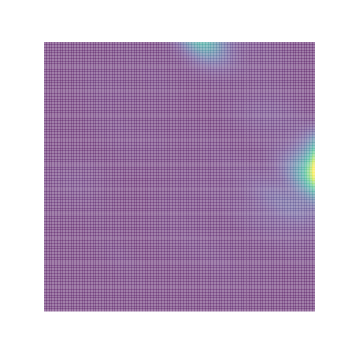}
&
\includegraphics[width=0.10\textwidth, trim=30px 25px 25px 30px, clip] {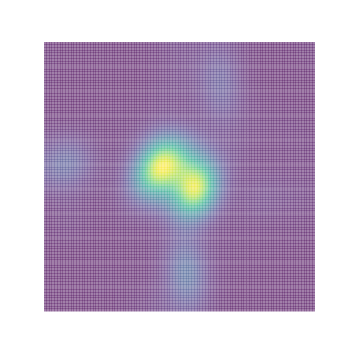}
\\\hline
\multicolumn{4}{c}{Livingroom. \ \ Top:Training, \ Middle: No-align,\  Bottom: Ours.} \\
Sofa-Table & Table-TV & Plant-Sofa & Sofa-TV \\
\includegraphics[width=0.10\textwidth, trim=30px 30px 30px 30px, clip]{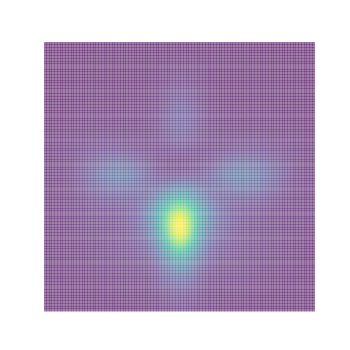}
&
\includegraphics[width=0.10\textwidth, trim=30px 30px 30px 30px, clip]{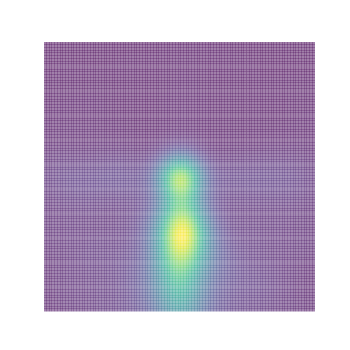}
&
\includegraphics[width=0.10\textwidth, trim=30px 30px 30px 30px, clip]{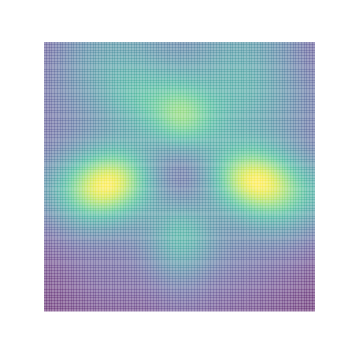}
&
\includegraphics[width=0.10\textwidth, trim=30px 30px 30px 30px, clip]{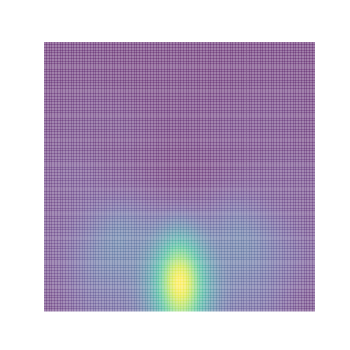}
\\
\includegraphics[width=0.10\textwidth, trim=30px 30px 30px 30px, clip]{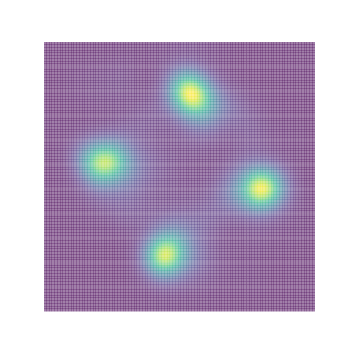}
&
\includegraphics[width=0.10\textwidth, trim=30px 30px 30px 30px, clip]{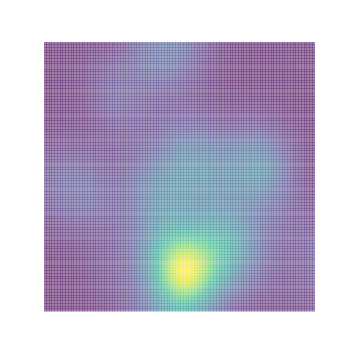}
&
\includegraphics[width=0.10\textwidth, trim=30px 30px 30px 30px, clip]{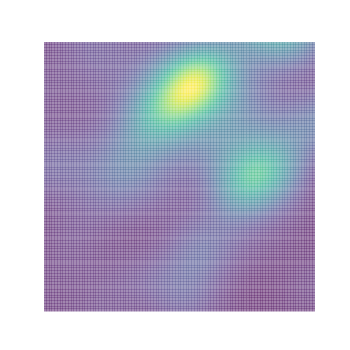}
&
\includegraphics[width=0.10\textwidth, trim=30px 30px 30px 30px, clip]{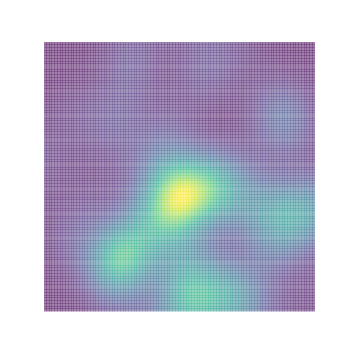}
\\
\includegraphics[width=0.10\textwidth, trim=30px 30px 30px 30px, clip]{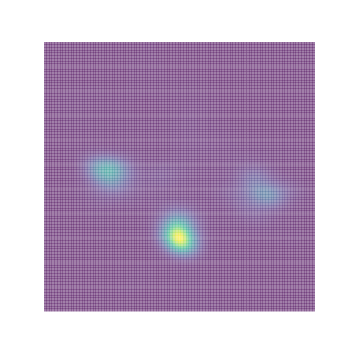}
&
\includegraphics[width=0.10\textwidth, trim=30px 30px 30px 30px, clip]{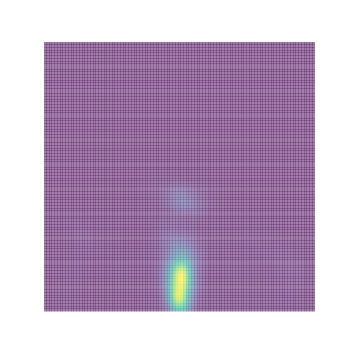}
&
\includegraphics[width=0.10\textwidth, trim=30px 30px 30px 30px, clip]{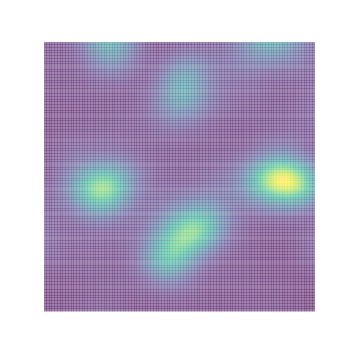}
&
\includegraphics[width=0.10\textwidth, trim=30px 30px 30px 30px, clip]{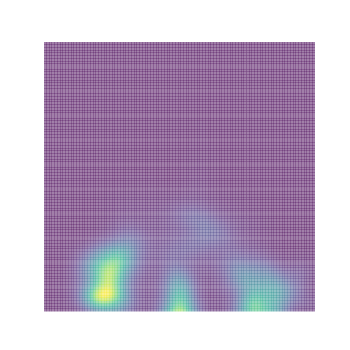}
\end{tabular}
\caption{Distributions of relative positions between selected object pairs. The origin lies in the image center. Top three rows: Distributions of selected pairs of classes in Bedroom for training data, synthesized scenes with no global scene alignment, and our synthesized scenes respectively (from left to right: Desk/Chair, Bed/Nightstand, Bed/Television, and Chair/Computer). Bottom three rows: Distributions of selected pairs of classes in Living Room for training data, synthesized scenes with no global scene alignment, and our synthesized scenes (from left to right: Sofa/Table, Table/Television, Plant/Sofa, and Sofa/Television).}
\label{Figure:Relative:Position:Distribution}
\end{figure}

\begin{figure}
\begin{tabular}{c|c|c|c}
\multicolumn{4}{c}{Bedroom. \ \ Top:Training, \ Middle: No-align,\  Bottom: Ours.} \\
Desk-Chair & Bed-Night. & Bed-TV. & Chair-Computer. \\
\includegraphics[width=0.10\textwidth, trim=30px 30px 30px 30px, clip]{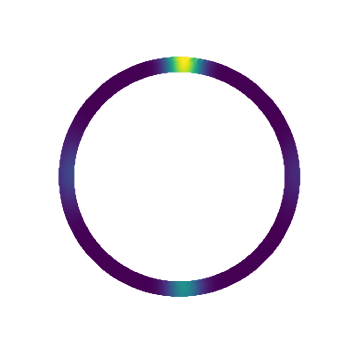}
&
\includegraphics[width=0.10\textwidth, trim=30px 30px 30px 30px, clip]{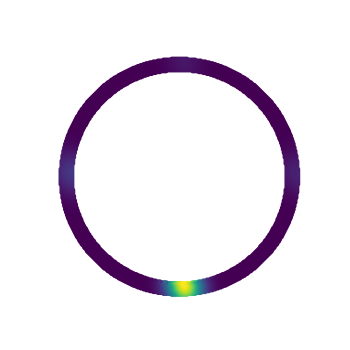}
&
\includegraphics[width=0.10\textwidth, trim=30px 30px 30px 30px, clip]{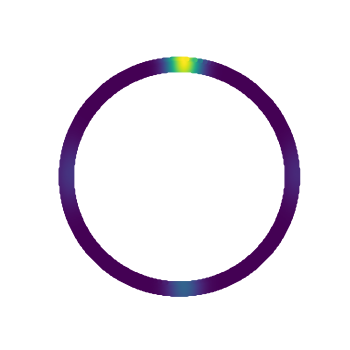}
&
\includegraphics[width=0.10\textwidth, trim=30px 30px 30px 30px, clip]{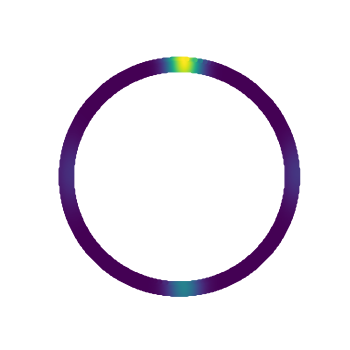}
\\
\includegraphics[width=0.10\textwidth, trim=30px 30px 30px 30px, clip]{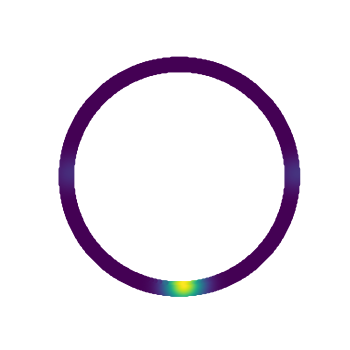}
&
\includegraphics[width=0.10\textwidth, trim=30px 30px 30px 30px, clip]{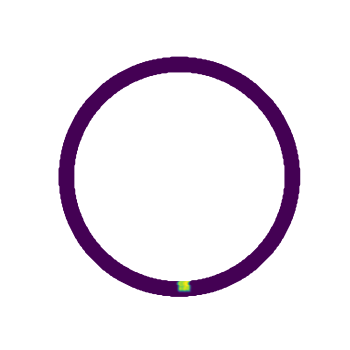}
&
\includegraphics[width=0.10\textwidth, trim=30px 30px 30px 30px, clip]{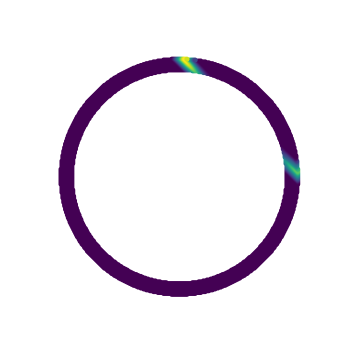}
&
\includegraphics[width=0.10\textwidth, trim=30px 30px 30px 30px, clip]{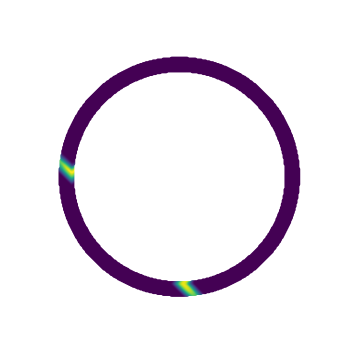}
\\
\includegraphics[width=0.10\textwidth, trim=30px 30px 30px 30px, clip]{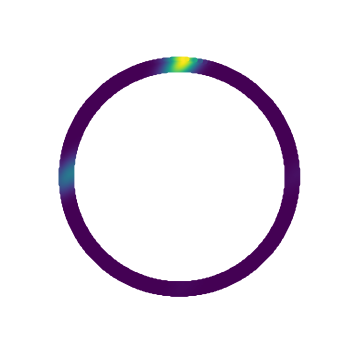}
&
\includegraphics[width=0.10\textwidth, trim=30px 30px 30px 30px, clip]{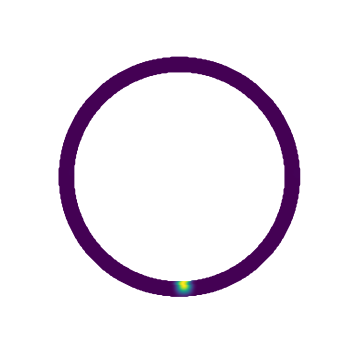}
&
\includegraphics[width=0.10\textwidth, trim=30px 30px 30px 30px, clip]{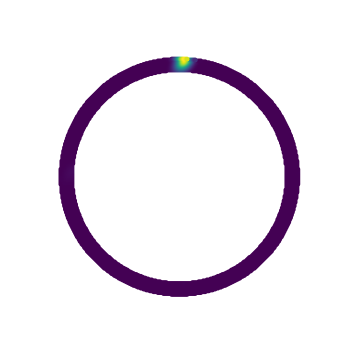}
&
\includegraphics[width=0.10\textwidth, trim=30px 30px 30px 30px, clip]{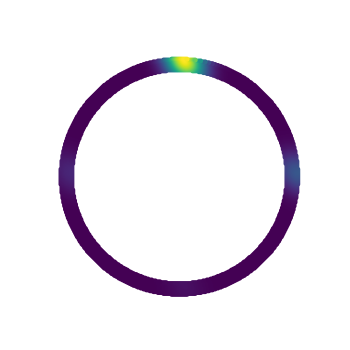}
\\\hline
\multicolumn{4}{c}{Livingroom. \ \ Top:Training, \ Middle: No-align,\  Bottom: Ours.} \\
Sofa-Table & Table-TV & Plant-Sofa & Sofa-TV \\
\includegraphics[width=0.10\textwidth, trim=30px 30px 30px 30px, clip]{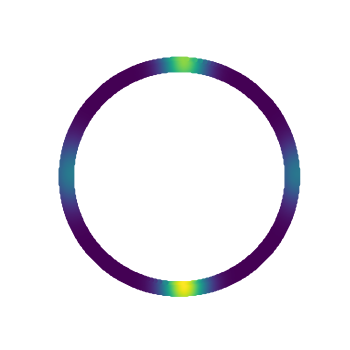}
&
\includegraphics[width=0.10\textwidth, trim=30px 30px 30px 30px, clip]{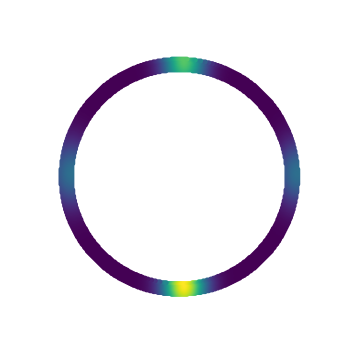}
&
\includegraphics[width=0.10\textwidth, trim=30px 30px 30px 30px, clip]{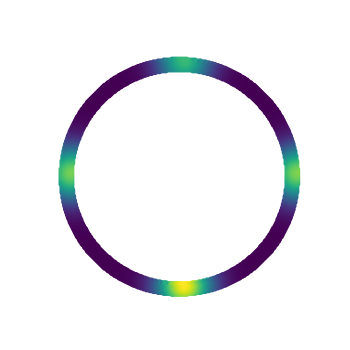}
&
\includegraphics[width=0.10\textwidth, trim=30px 30px 30px 30px, clip]{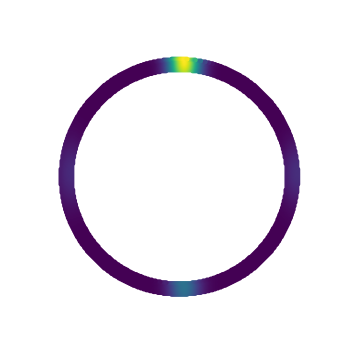}
\\
\includegraphics[width=0.10\textwidth, trim=30px 30px 30px 30px, clip]{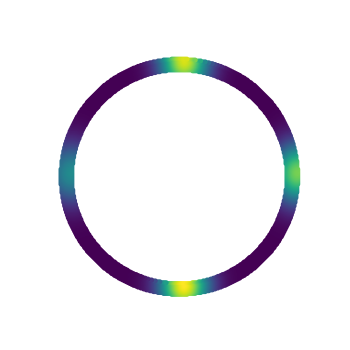}
&
\includegraphics[width=0.10\textwidth, trim=30px 30px 30px 30px, clip]{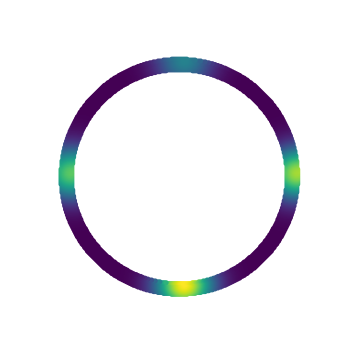}
&
\includegraphics[width=0.10\textwidth, trim=30px 30px 30px 30px, clip]{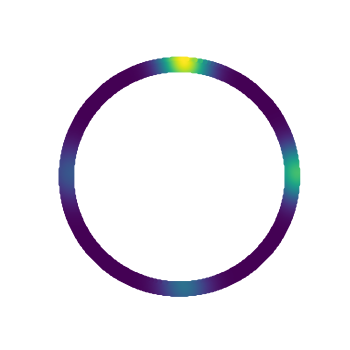}
&
\includegraphics[width=0.10\textwidth, trim=30px 30px 30px 30px, clip]{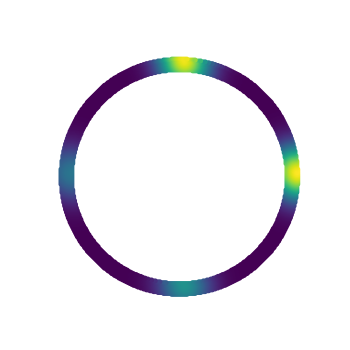}
\\
\includegraphics[width=0.10\textwidth, trim=30px 30px 30px 30px, clip]{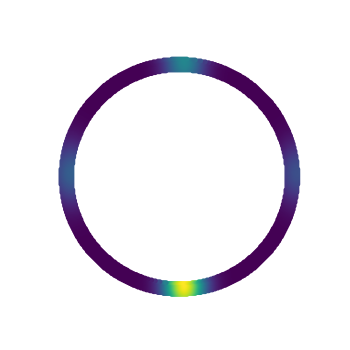}
&
\includegraphics[width=0.10\textwidth, trim=30px 30px 30px 30px, clip]{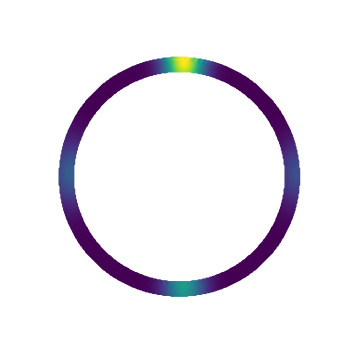}
&
\includegraphics[width=0.10\textwidth, trim=30px 30px 30px 30px, clip]{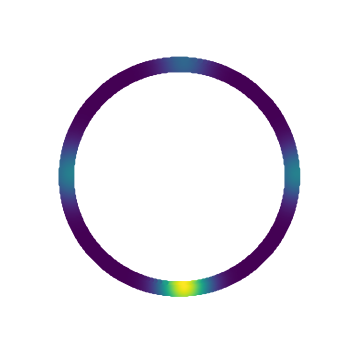}
&
\includegraphics[width=0.10\textwidth, trim=30px 30px 30px 30px, clip]{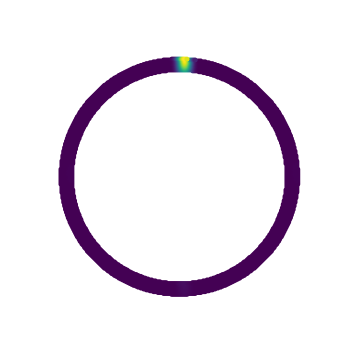}
\end{tabular}
\caption{Distributions of relative orientations between selected object pairs. The bottom of the circle represents the case where the two objects share the same orientation. Top three rows: Distributions of selected pairs of classes in Bedroom for training data, synthesized scenes with no global scene alignment, and our synthesized scenes respectively (from left to right: Desk/Chair, Bed/Nightstand, Bed/Television, and Chair/Computer). Bottom three rows: Distributions of selected pairs of classes in Living Room for training data, synthesized scenes with no global scene alignment, and our synthesized scenes (from left to right: Sofa-Table, Table-Television, Plant/Sofa, and Sofa/Television).}
\label{Figure:Angular:Correlation:Distribution}
\end{figure}

As illustrated in Figure~\ref{Figure:Relative:Position:Distribution}, our generator nicely matches the distributions in the training data. In particular, for Desk/Chair and Bed/Nightstand, the learned distribution and the original distribution are very similar to each other. In other words, our approach learns important pairwise relations in the training data. Figure ~\ref{Figure:Angular:Correlation:Distribution} shows the distribution of the relative angles between the front orientations. We have quantized the generated angles, range from $0$ to $2\pi$, into 4 bins. The bottom of the circle corresponds to the case where the two object shares the same orientation. Again, our method learns such distributions reasonably well.

\subsection{The Importance of Joint Scene Alignment}
\label{Section:Ablation:Study}

In this section, we perform an additional study to justify that jointly optimizing 3D scenes as a preprocessing step is important. As an evaluation metric, we use the distribution between selected pairs of objects between the Chair class and Table class and the Bed class and the NightStand class locations and orientation on the Bedroom dataset.


\paragraph{Global scene alignment} As illustrated in Figure~\ref{Figure:Relative:Position:Distribution} and ~\ref{Figure:Angular:Correlation:Distribution}, deactivating the global scene alignment step (i.e., applying our alternating minimization procedure on the raw input data directly) causes the network to not learn correlations between salient patterns. 
The distributions of relative locations on generated scenes are significantly different from that on the training data. This justifies that global scene alignment is crucial to the success of our system. In other words, it is insufficient for local formulation to align the input scenes.  



\subsection{Applications in Scene Interpolation}
\label{Section:Scene:Interpolation}

\begin{figure*}
\includegraphics[width=0.16\textwidth, trim=25px 45px 25px 0px, clip]{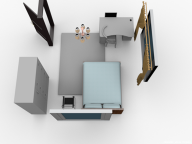}
\includegraphics[width=0.16\textwidth, trim=25px 45px 25px 0px, clip]{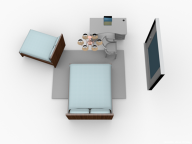}
\includegraphics[width=0.16\textwidth, trim=25px 45px 25px 0px, clip]{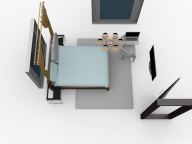}
\includegraphics[width=0.16\textwidth, trim=25px 45px 25px 0px, clip]{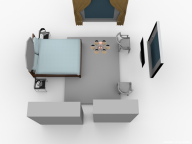}
\includegraphics[width=0.16\textwidth, trim=25px 45px 25px 0px, clip]{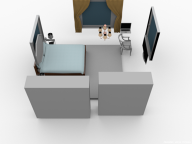}
\includegraphics[width=0.16\textwidth, trim=25px 45px 25px 0px, clip]{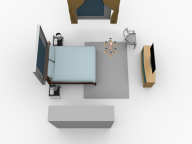}
\includegraphics[width=0.16\textwidth, trim=25px 45px 25px 0px, clip]{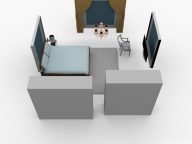}
\includegraphics[width=0.16\textwidth, trim=25px 45px 25px 0px, clip]{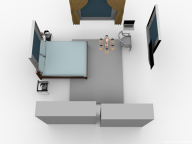}
\includegraphics[width=0.16\textwidth, trim=25px 45px 25px 0px, clip]{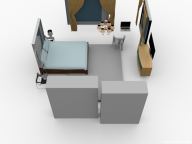}
\includegraphics[width=0.16\textwidth, trim=25px 45px 25px 0px, clip]{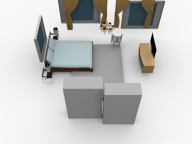}
\includegraphics[width=0.16\textwidth, trim=25px 45px 25px 0px, clip]{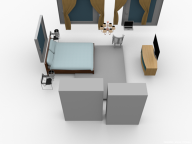}
\includegraphics[width=0.16\textwidth, trim=25px 45px 25px 0px, clip]{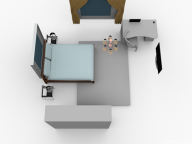}
\includegraphics[width=0.16\textwidth, trim=25px 45px 25px 0px, clip]{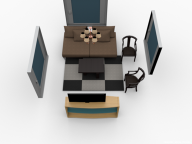}
\includegraphics[width=0.16\textwidth, trim=25px 45px 25px 0px, clip]{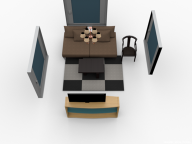}
\includegraphics[width=0.16\textwidth, trim=25px 45px 25px 0px, clip]{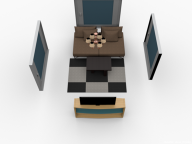}
\includegraphics[width=0.16\textwidth, trim=25px 45px 25px 0px, clip]{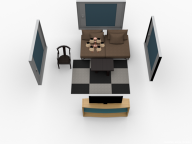}
\includegraphics[width=0.16\textwidth, trim=25px 45px 25px 0px, clip]{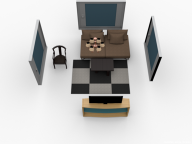}
\includegraphics[width=0.16\textwidth, trim=25px 45px 25px 0px, clip]{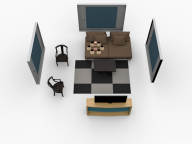}
\includegraphics[width=0.16\textwidth, trim=25px 45px 25px 0px, clip]{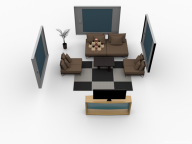}
\includegraphics[width=0.16\textwidth, trim=25px 45px 25px 0px, clip]{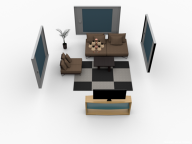}
\includegraphics[width=0.16\textwidth, trim=25px 45px 25px 0px, clip]{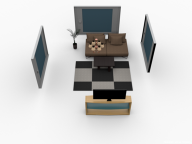}
\includegraphics[width=0.16\textwidth, trim=25px 45px 25px 0px, clip]{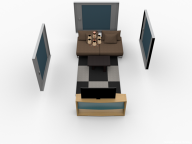}
\includegraphics[width=0.16\textwidth, trim=25px 45px 25px 0px, clip]{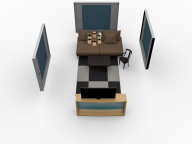}
\includegraphics[width=0.16\textwidth, trim=25px 45px 25px 0px, clip]{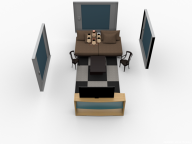}
\caption{Scene interpolation results between different pairs of source (left column) and target (right column) scenes.}
\label{Figure:Application:Scene:Interpolation}
\end{figure*}

In this section, we show the application of our approach in scene interpolation. Given two input scenes, we first compute their associated latent parameters. We then interpolate these two latent parameters along the straight line between the two parameters and use the generator to generate the corresponding synthetic scenes. Figure~\ref{Figure:Application:Scene:Interpolation} shows four examples. The first two examples are interpolations of bedroom scenes, and the second two are from living room scenes. For bedroom scenes, the first example consists of two scenes with very different configurations, where the orientation of two rooms are not aligned. The intermediate scenes gradually remove the original bed, table and desk, and then add the bed with new location and orientation, which is semantically meaningful. The second example consists of two similar bedroom scenes with different objects. In intermediate scenes, bed, night stands and table lamps stay the same, while wardrobes are gradually removed and the a table and a laptop are added gradually. These are meaningful interpolations. For living room scenes, the first example consists of two scenes with similar configuration of objects but the locations of the chairs are reversed with respect to the sofa object. The intermediate scenes gradually remove objects on one side and then add objects on the other side, leading to a meaningful interpolation. The second example shows a configuration where the source scene has different objects than the target scene. The intermediate scenes progressively delete objects and then add new objects in different category, which is again semantically meaningful. 

\subsection{Applications in Scene Completion}
\label{Section:Scene:Completion}

\begin{figure*}
\includegraphics[width=0.16\textwidth, trim=25px 30px 25px 0px, clip]{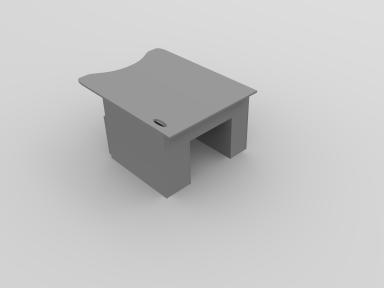}
\includegraphics[width=0.16\textwidth, trim=25px 30px 25px 0px, clip]{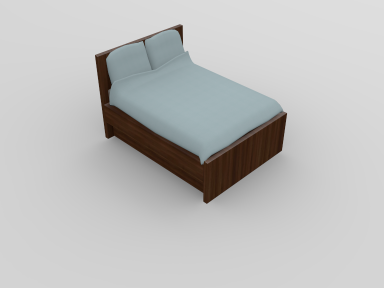}
\includegraphics[width=0.16\textwidth, trim=25px 30px 25px 0px, clip]{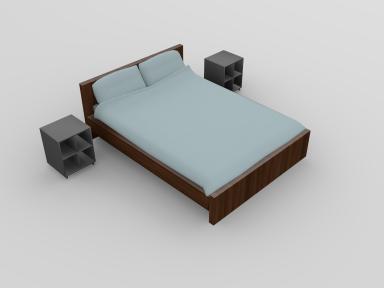}
\includegraphics[width=0.16\textwidth, trim=25px 30px 25px 0px, clip]{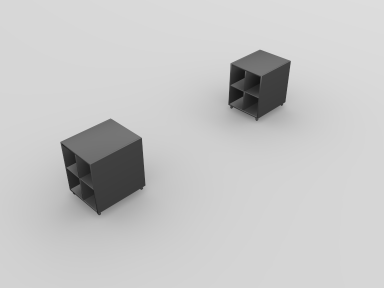}
\includegraphics[width=0.16\textwidth, trim=25px 30px 25px 0px, clip]{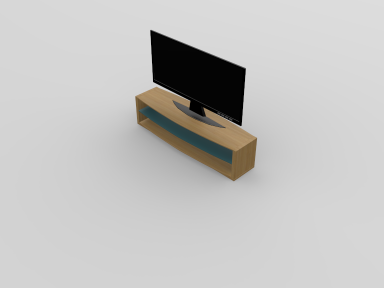}
\includegraphics[width=0.16\textwidth, trim=25px 30px 25px 0px, clip]{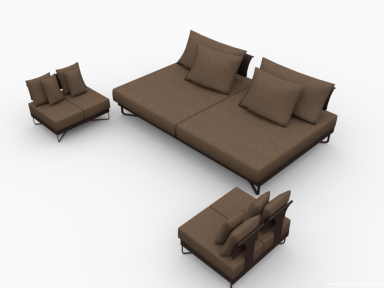}

\includegraphics[width=0.16\textwidth, trim=15px 30px 15px 0px, clip]{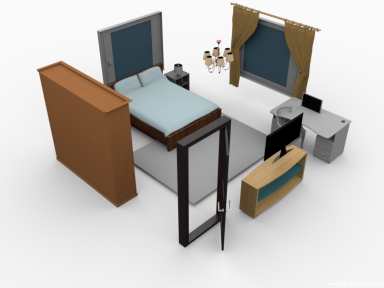}
\includegraphics[width=0.16\textwidth, trim=15px 30px 15px 0px, clip]{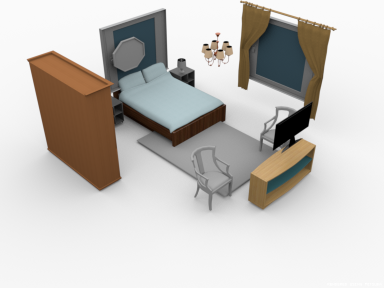}
\includegraphics[width=0.16\textwidth, trim=15px 30px 15px 0px, clip]{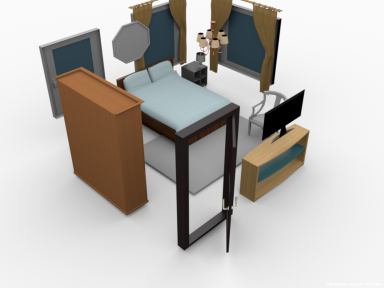}
\includegraphics[width=0.16\textwidth, trim=15px 30px 15px 0px, clip]{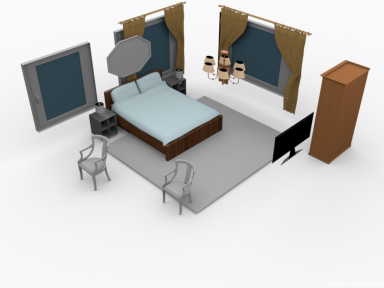}
\includegraphics[width=0.16\textwidth, trim=15px 30px 15px 0px, clip]{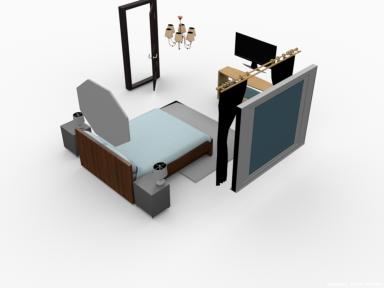}
\includegraphics[width=0.16\textwidth, trim=15px 30px 15px 0px, clip]{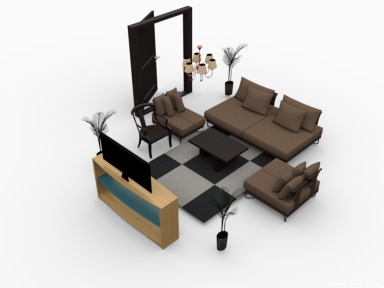}

\includegraphics[width=0.16\textwidth, trim=15px 30px 15px 0px, clip]{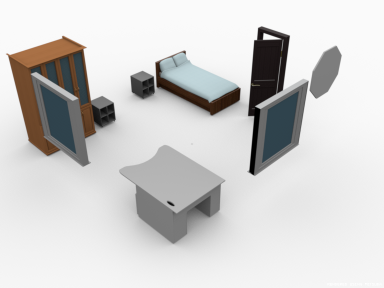}
\includegraphics[width=0.16\textwidth, trim=15px 30px 15px 0px, clip]{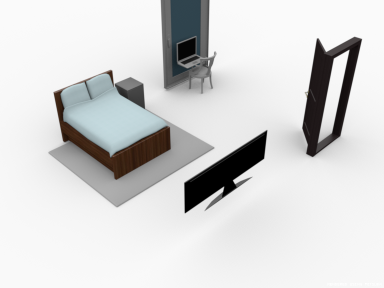}
\includegraphics[width=0.16\textwidth, trim=15px 30px 15px 0px, clip]{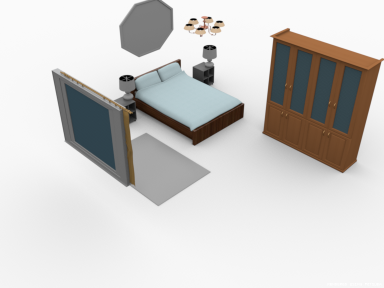}
\includegraphics[width=0.16\textwidth, trim=15px 0px 15px 30px, clip]{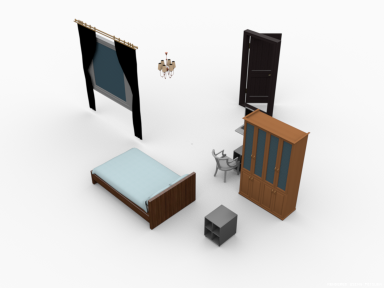}
\includegraphics[width=0.16\textwidth, trim=15px 30px 15px 0px, clip]{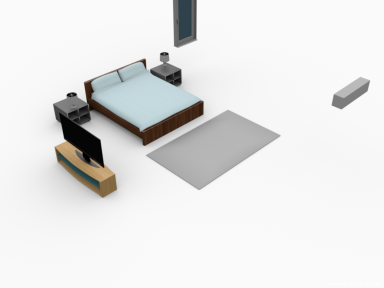}
\includegraphics[width=0.16\textwidth, trim=15px 15px 15px 15px, clip]{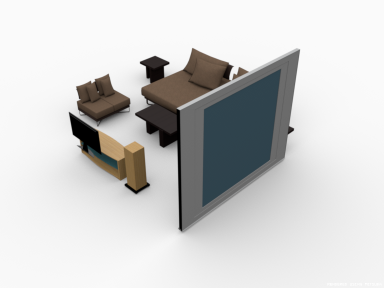}

\includegraphics[width=0.16\textwidth, trim=0px 0px 0px 0px, clip]{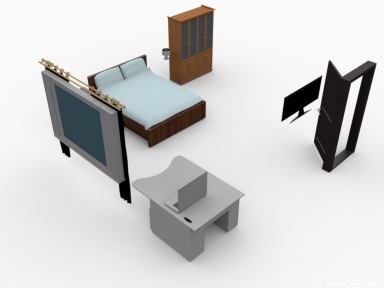}
\includegraphics[width=0.16\textwidth, trim=0px 0px 0px 0px, clip]{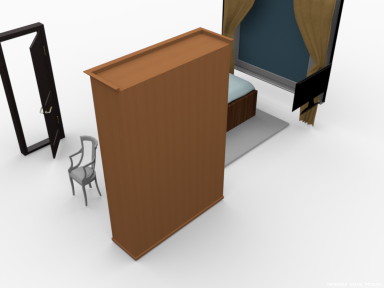}
\includegraphics[width=0.16\textwidth, trim=0px 0px 0px 0px, clip]{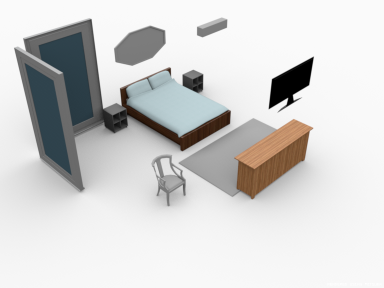}
\includegraphics[width=0.16\textwidth, trim=0px 0px 0px 0px, clip]{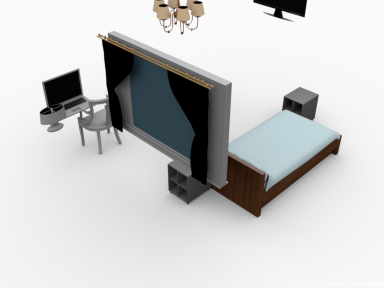}
\includegraphics[width=0.16\textwidth, trim=0px 0px 0px 0px, clip]{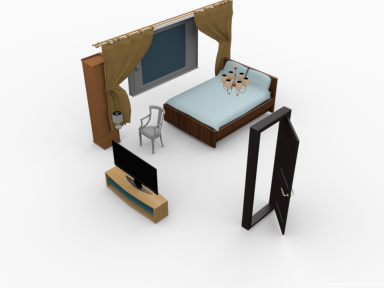}
\includegraphics[width=0.16\textwidth, trim=0px 0px 0px 0px, clip]{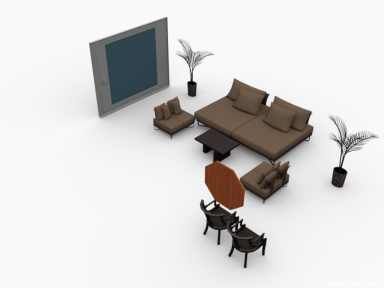}
\caption{Scene completion results. From top to bottom, we show the input objects, completed scenes generated by our method, and the results using~\protect\cite{Fisher:2012:ESO} and~\protect\cite{Kermani:2016:LSS}, respectively.}
\label{Figure:Application:Scene:Completion}
\end{figure*}

We then show the application of our approach on the application of scene completion. In this task, we are given a partial scene, and our task is to find the optimal scene that completes the input scene. Towards this end, we solve the following optimization problem:
\begin{equation}
\bs{z}^{\star} = \underset{\bs{z}, \set{T}, \set{S}}{\textup{argmin}} \ \|\set{T}\set{S}(M_{in}) - C.G_{\theta}(\bs{z})\|_F^2 + \alpha \|\bs{z}\|^2
\label{Eq:Scene:Completion}
\end{equation}
where $C$ is the mask associated with $M_{in}$, and it constraints that the completed scene should contain objects in the input partial scene. $.$ is the elementwise matrix product. Again we apply gradient descent for optimization. We set $\alpha = 1e-3$ in our experiments.

As a comparison, we compare the synthesis results of \cite{Fisher:2012:ESO} and \cite{Kermani:2016:LSS}. Since both approaches used different datasets, as for a fair comparison, we reimplemented their approaches on our Bedroom and Living Room datasets. 

Figure~\ref{Figure:Application:Scene:Completion} compares our approach with baseline approaches. The input partial scenes are cropped from scenes in the testing datasets. Since both baseline approaches generate a series of completed scenes, for a fair comparison we choose the ones that have the same number of objects as the output of our approach. We can see that our approach leads to semantically more meaningful results in terms of both groups of co-related objects  and locally compatible of object pairs. We can understand this as the fact that our approach optimizes the scene layout with respect to all patterns captured by the neural network. In contrast, both baseline approaches are sequential, despite the usage of local optimization~\cite{Yu:2011:MHA} to improve scene layouts, they may not be able to explore the entire underlying scene space for generating the completed scenes. 

Moreover, our approach is significantly faster than the baseline approaches. Our approach takes 1-2 seconds for solving (\ref{Eq:Scene:Completion}) , while \cite{Fisher:2012:ESO} and \cite{Kermani:2016:LSS} take 83.1 seconds and 76.4 seconds in average, respectively. In particular, most of the computational time was spent on running local optimization to improve the scene layouts. 
\section{Conclusions}
\label{Section:Conclusions}

We have studied the problem of 3D scene synthesis using deep generative models. Unlike 2D images, 3D geometries possess multiple varying representations, each with its advantages and disadvantages for the most efficacious deep neural networks. To maximize tradeoffs, we therefore presented a hybrid methodology that trains a 3D scene generator using  a combination of a 3D object arrangement representation, and a projected 2D image representation. combine the advantages of both representations. The, 3D object arrangement representation ensures local and global neighborhood structure of the synthesized scene, while image-based representations preserve local view dependent patterns. Moreover the results obtained from the image-based representation is beneficial for training the 3D generator.

Our 3D scene generator is a feed-forward neural network. This network design takes another route from the common recurrent methodology for 3D scene synthesis and modeling. The benefit of the feed-forward architecture is that it can jointly optimize all the factors for 3D synthesis, while it is difficult for a recurrent approach to recover from mistakes made during its sequential processing. Preliminary qualitative evaluations have shown the advantage of the feed-forward architecture over two recurrent approaches. Although it is premature to say that feed-forward approaches shall significantly dominate recurrent approaches, we do believe that free-forward networks have shown great promise in several scenarios, and deserves further research and exploitation.   

One limitation of our approach is that we do not completely encode physical properties of the synthesized scenes, which are important for computer aided manufacturing purposes (e.g., 3D printing) . To address this issue, one possibility is to develop a suitable 3D representation that explicitly encodes physical properties, e.g., using a shape grammar.

Another limitation of our approach is that all the training data should consist of semantically segmented 3D scenes. This may not always be possible, e.g., reconstructed 3D scenes from point clouds are typically not segmented into individual objects. A potential way to address this issue is to extend the consistent hybrid representation described in this paper, e.g., by enforcing the consistency among three networks: 1) scene synthesis under the image-based representation, 2) scene synthesis under the 3D object arrangement representation and 3) a network that converts a 3D scene into its corresponding 3D object arrangement representation. 

There are multiple directions and opportunities for future research. As mentioned in the introduction, there are at least five frequently used 3D representations. One could  extend our current approach to use more than one 3D representation. For example, we could leverage multi-view representations on which we have rich training data (e.g., internet images). The multi-view representation also provides texture information, useful for synthesizing 3D representations. Finally, we would propose to combine the learned 3D representation with data from other modalities such as natural language descriptions.

\noindent\textbf{Acknowledgement.} The authors would like thank Chandrajit Bajaj for many fruitful discussions. Qixing Huang would like to acknowledge support of this research from NSF DMS-1700234, a Gift from Snap Research, and a hardware Donation from NVIDIA.

\appendix
\section{Additional Details on Pair-wise Scene Alignment}
\label{Eq:Pairwise:Scene:Alignment} 

In this section, we present our numerical optimization approach for solving (\ref{Eq:Pairwise:Alignment:Objective}), which combines reweighted non-linear least squares and alternating minimization. To this end, we first introduce a weight vector $\bs{w}$ corresponding to the columns of $\set{T}(\set{S}(M_i))-M_j$ and modify the optimization problem as 
\begin{equation}
\set{T}_{ij}^{\init}, \set{S}_{ij}^{\init} = \underset{\set{T}, \set{S}}{\textup{argmin}}\ \|(\set{T}(\set{S}(M_i))-M_j)\diag(\bs{w})\|_{F}^2.
\label{Eq:Pairwise:Alignment:Objective2}
\end{equation}
RLSM alternates between fixing $\bs{w}$ to solve (\ref{Eq:Pairwise:Alignment:Objective2}) and using the optimal solution to update $\bs{w}$. We set the initial weight vector as $\bs{w}^{(0)} = \bs{1}$.

Given the weight vector $\bs{w}^{(t)}$ at iteration $t$, we again perform alternating minimization to optimize $\set{T}$ and $\set{S}$. At each inner iteration $s$, the updates are given by
\begin{align}
\set{T}^{(t,s)} = \underset{\set{T}}{\textup{argmin}}\ \|(\set{T}(\set{S}^{(t,s-1)}(M_i))-M_j)\diag(\bs{w}^{(t)})\|_{F}^2, \label{Eq:T:opt} \\
\set{S}^{(t,s)} = \underset{\set{S}}{\textup{argmin}}\ \|(\set{T}^{(t,s)}(\set{S}(M_i))-M_j)\diag(\bs{w}^{(t)})\|_{F}^2. \label{Eq:S:opt}
\end{align}
In this case, both (\ref{Eq:T:opt}) and (\ref{Eq:S:opt}) admit closed-form solutions. The optimal solution of (\ref{Eq:T:opt}) can be computed using \cite{Horn87closed-formsolution}. The optimal solution of (\ref{Eq:S:opt}) can be computed by solving a linear assignment. 

This alternate minimization procedure converges fairly fast, we apply 4 iterations in our implementation.

Given the solution $\set{T}^{(t)}$ and $\set{S}^{(t)}$ from the alternating minimization procedure described above, we update the weight vector at iteration $t+1$ as
$$
w_k^{(t+1)} = \epsilon\slash\sqrt{\epsilon^2 + \|(\set{T}^{(t)}\set{S}^{(t)}(M_i)-M_j)\bs{e}_k\|^2}, \quad 1\leq k \leq K,
$$
where $\bs{e}_k$ is the k-th canonical basis of $\R^{K}$. $\epsilon=10^{-3}$ is chosen to be a small value. In our implementation, we apply 4 iterations of reweighted least squares.


\section{Gradient of the Image Projection}
\label{Section:Projection:Operator}

Since $\set{P}(M)$ is an image, and the pixel values are summation of signed distance function values. In addition, the signed distance function is with respect to a oriented box. Thus, it is sufficient to derive the formula for computing the gradient of a point $\bs{p}$ with respect to a line $l$ parameterized by an orientation $\bs{n}$ and a point $\bs{q} = \bs{o} + s\bs{n}$ on $l$:
\begin{align}
d(\bs{p},l) &:= (\bs{p} - \bs{q})^{T}\bs{n} \nonumber \\
& = (\bs{p} - \bs{o} - s\bs{n})^{T}\bs{n} \nonumber \\
& = (\bs{p} - \bs{o})^{T}\bs{n} - s.
\end{align}
Here $\bs{o}$ represents the center of the box, $\bs{n}$ is the axis that is perpendicular to the line, and $s$ is the size along this axis. 

It is easy to see that the derivative of $d(\bs{p},l)$ with respect to $\bs{o}$, $\bs{n}$ and $s$ are given by
\begin{equation}
\frac{\partial d(\bs{p},l)}{\partial \bs{o}} = -\bs{n},\quad 
\frac{\partial d(\bs{p},l)}{\partial \bs{n}} = \big((\bs{p} - \bs{o})^{T}\bs{n}^{\perp}\big)\cdot\bs{n}^{\perp}, \quad 
\frac{\partial d(\bs{p},l)}{\partial s} = -1.
\end{equation}
Here $\bs{n}^{\perp}$ is a vector that is perpendicular to $\bs{n}$.
\section{Statistics on SUNCG}
\label{Section:Statistics:SUNCG}

Table~\ref{Table:Bedroom} and Table~\ref{Table:Livingroom} collect statistics on Bedroom and Living Room, respectively.

\begin{table}[t]
\begin{tabular}{|c|c|c|c|}\hline
Name     & window & bed &  wardrobe \\
Count    &8156    &    7288    &    7134 \\ \hline
Name     & stand &  door & table lamp \\
Count    &       6921    &   6715     &  6375  \\ \hline
Name     & television &  curtain & rug\\
Count    &5396    &   5111     &  4705  \\ \hline
Name     & computer & computer  & chandelier\\
Count    &4687    &   4551     &  4372 \\ \hline
Name     & desk& picture frame & shelving\\
Count    &4246    &   3772     &  3674  \\ \hline
Name     & dresser & plant &  table \\
Count    & 3333    &   3168     &  2647  \\\hline
Name     & dressing table & tv stand & books\\
Count    & 2473    &   2433     &  2339  \\\hline
Name      & ottoman&  mirror & air conditioner \\
Count    & 2256    &   2155     &  2153 \\\hline
Name     & floor lamp & wall lamp & sofa \\
Count    & 2050    &   1953     &  1651 \\\hline
Name     & vase & hanger & heater \\
Count    &       1642    &   1182     &  1104 \\\hline
\end{tabular}
\caption{Names of classes and number of instances in each class of the Bedroom dataset.}
\label{Table:Bedroom}
\end{table}

\begin{table}[t]
\begin{tabular}{|c|c|c|c|}\hline
Name     & sofa & window &  table \\
Count    &   9608   &     8217  &      7873  \\ \hline
Name     & chair &  television & plant \\
Count    &  5898    &   5680     &   5070  \\ \hline
Name     & door &  chandelier & rug\\
Count    &  4480    &   4129     &  3978  \\ \hline
Name     & curtain & tv stand  & picture frame\\
Count    &  3936    &   3778     &  3385 \\ \hline
Name     & shelving& floor lamp & loudspeaker\\
Count    &  3082    &   3059     &  2599  \\ \hline
Name     & vase & ottoman &  computer \\
Count     &  2514    &   1871     &  1841  \\\hline
Name     & books & fireplace & air conditioner\\
Count    &  1773    &   1389     &  1341 \\\hline
Name      & wall lamp&  wardrobe & clock \\
Count    &  1286    &   1255     &  1238 \\\hline
Name     & stereo set & kitchen cabinet & desk \\
Count    &  1204    &   1178     &  1159 \\\hline
Name     & heater & fish tank & playstation \\
Count    &  1016    &    936     &   906 \\\hline
\end{tabular}
\caption{Names of classes and number of instances in each class of the Living Room dataset.}
\label{Table:Livingroom}
\end{table}

\end{document}